\documentclass{article}

\PassOptionsToPackage{square,numbers}{natbib}

\usepackage[final]{neurips_2025}

\usepackage[utf8]{inputenc} 
\usepackage[T1]{fontenc}    
\usepackage{hyperref}       
\usepackage{url}            
\usepackage{booktabs}       
\usepackage{amsfonts}       
\usepackage{nicefrac}       
\usepackage{microtype}      
\usepackage{xcolor}         
\usepackage{graphicx}       
\usepackage{amsmath}        
\usepackage{longtable}

\title{Predictive Coding Enhances Meta-RL To Achieve Interpretable Bayes-Optimal Belief Representation Under Partial Observability}

\author{
  Po-Chen Kuo\\
  University of Washington \\
  \texttt{pckuo@uw.edu}
  \And
  Han Hou\\
  Allen Institute for Neural Dynamics \\
  \texttt{han.hou@alleninstitute.org}
  \AND
  Will Dabney\\
  Google DeepMind \\
  \texttt{wdabney@gmail.com}
  \And
  Edgar Y. Walker\\
  University of Washington \\
  \texttt{eywalker@uw.edu}
}

\begin{document}

\maketitle

\begin{abstract}

Learning a compact representation of history is critical for planning and generalization in partially observable environments. 
While meta-reinforcement learning (RL) agents can attain near Bayes-optimal policies, they often fail to learn the compact, interpretable Bayes-optimal belief states. 
This representational inefficiency potentially limits the agent's adaptability and generalization capacity.
Inspired by predictive coding in neuroscience---which suggests that the brain predicts sensory inputs as a neural implementation of Bayesian inference---and by auxiliary predictive objectives in deep RL, we investigate whether integrating self-supervised predictive coding modules into meta-RL can facilitate learning of Bayes-optimal representations.
Through state machine simulation, we show that meta-RL with predictive modules consistently generates more interpretable representations that better approximate Bayes-optimal belief states compared to conventional meta-RL across a wide variety of tasks, even when both achieve optimal policies.
In challenging tasks requiring active information seeking, only meta-RL with predictive modules successfully learns optimal representations and policies, whereas conventional meta-RL struggles with inadequate representation learning.
Finally, we demonstrate that better representation learning leads to improved generalization.
Our results strongly suggest the role of predictive learning as a guiding principle for effective representation learning in agents navigating partial observability.

\end{abstract}

\section{Introduction}

In real-world environments, agents, biological or artificial, rarely have complete information about the environmental states crucial for decision-making and planning, as observations are often noisy and non-stationary. 
This issue is known in reinforcement learning (RL) as partial observability \cite{Sutton1998, cassandra1994acting, kaelbling1998planning}, and is a major challenge in the deployment of real-world RL systems \cite{dulac2021challenges}. 
In partially observable environments, learning optimal policies depends on the entire history of past interactions. 
An open question in RL under partial observability is how to learn an efficient representation that serves as a compact summary of the history while retaining the information crucial for policy learning.

Partially observable tasks can be formalized as partially observable Markov Decision Processes (POMDPs, Fig. \ref{fig:intro}A) \cite{cassandra1994acting, kaelbling1998planning}, allowing a Bayesian treatment by maintaining and updating a belief state using Bayesian inference \cite{subramanian2022approximate}. 
To this end, meta-RL, in particular memory-based meta-RL, has been shown to provide powerful deep RL methods for developing agents that efficiently adapt under uncertainty \cite{duan2016rl, wang2016learning, igl2018deep, zintgraf2021varibad, akuzawa2021estimating, beck2023survey}. 
Meta-RL has been shown to behave near Bayes-optimally under partial observability, both theoretically and empirically \cite{ortega2019meta, mikulik2020meta}.
However, its learned representations are not equivalent to the minimally sufficient, Bayes-optimal belief states \cite{mikulik2020meta}, hindering its interpretability, robustness, generalization capability, and learning of complex exploration \cite{akuzawa2021estimating, beck2023survey, mikulik2020meta}.

Predictive learning has been suggested as a central coding principle that guides efficient neural processing in neuroscience.
Humans and animals are shown to achieve near Bayes-optimal behavior under uncertainty, leading to theories and experiments investigating efficient Bayes-optimal representation and computation in the brain \cite{ernst2002humans, ma2006bayesian, kording2004bayesian}.
Predictive coding, postulating that the brain continuously predicts upcoming observations and updates its internal world models \cite{rao1999predictive, friston2005theory}, provides a neurally plausible implementation of Bayesian inference \cite{summerfield2014expectation, rao2010decision} and has been instrumental in explaining diverse neural circuits including feature learning in sensory cortices \cite{furutachi2024cooperative, zhuang2021unsupervised}, motor control in the cerebellum \cite{wolpert1998internal}, value learning in the striatum \cite{schultz1997neural}, and cognitive maps in the hippocampus \cite{o1978hippocampus}.
In deep learning, predictive learning has been explored as a powerful self-supervised technique to support representation learning and improve task performance \cite{oord2018representation, jaderberg2016reinforcement}.
In deep RL, predictive objectives are shown to regularize learning and prevent overfit or representational collapse \cite{lyle2021effect, dabney2021value}, improve sample efficiency \cite{hafner2019learning}, and link to Bayesian filtering \cite{lim2020recurrent}. 
Most recent breakthrough employs predictive learning to derive general agents that can solve a wide range of challenging tasks \cite{hafner2025mastering}.

The main contribution of this paper is a systematic investigation of whether integrating self-supervised predictive coding into meta-RL yields interpretable, Bayes-optimal belief representations in partially observable environments.
Although recent meta-RL models have explored predictive objectives to improve generalization and representation quality \cite{zintgraf2021varibad, akuzawa2021estimating}, the exact mechanisms by which predictive learning enhances latent representations---and how this relates to improved performance---remain unclear.
To fill this gap, we move beyond performance measures that lack interpretability to directly compare meta-RL agents against Bayes-optimal solutions, evaluating both behavior and representation across diverse partially observable tasks varying in belief-update complexity.
Specifically, we:

\begin{itemize}
    \item Employ a meta-RL framework incorporating self-supervised future predictive modules, which enables direct interpretation and comparison with Bayes-optimal belief states.
    \item Show systematically that meta-RL with predictive modules yields interpretable, task-relevant representations that capture underlying environmental structures and dynamics across diverse POMDPs, and significantly improves planning, exploration, and generalization. 
    \item Demonstrate via rigorous state machine simulation analysis that meta-RL with predictive modules consistently generates representations that more closely approximate Bayes-optimal beliefs than conventional black-box meta-RL, providing a guiding principle for learning both optimal representations and policies in partially observable environments.
\end{itemize}

\section{Background and related work}

\paragraph{Partially observable Markov decision process (POMDP)}
A POMDP is defined by the tuple $\bigl(S, A, T, R, \Omega, O, \gamma\bigr)$, where $S$ is the (finite) hidden state space, $A$ the action space, $T(s' \mid s,a)$ the transition probabilities, $R(s,a)$ the reward function, $\Omega$ the observation space, $O(o\mid s',a)$ the emission function, and $\gamma\in[0,1)$ the discount factor.  
Since the agent does not have access to the true state $s$, the decision process is non-Markovian with respect to $s$ and an optimal POMDP policy generally depends on the entire history $\tau_{:t}=(o_1,\dots,a_{t-1},o_t)$.
By maintaining a belief state $b_t\equiv p(s | h_t)$, a posterior over $S$ summarizing the history, the POMDP can be cast as a fully-observable \emph{belief} MDP with transitions in belief state $b$ given by the Bayesian update: $b'(s') = \eta\,O\bigl(o\mid s',a\bigr)\sum_{s\in S}T\bigl(s'\mid s,a\bigr)\,b(s)$, where $\eta$ is a normalization factor.  
In this formulation, belief states $b$ are the minimally sufficient statistic of the history summarizing all the past information compactly, and the policy $\pi(b)$ over belief states restores the Markov property.
Unfortunately, planning in a POMDP, i.e., computing a Bayes-optimal policy over belief states, is generally intractable for all but the simplest tasks \cite{cassandra1994acting, kaelbling1998planning, subramanian2022approximate}.

\begin{figure}
  \centering
  \includegraphics[width=1.0\textwidth]{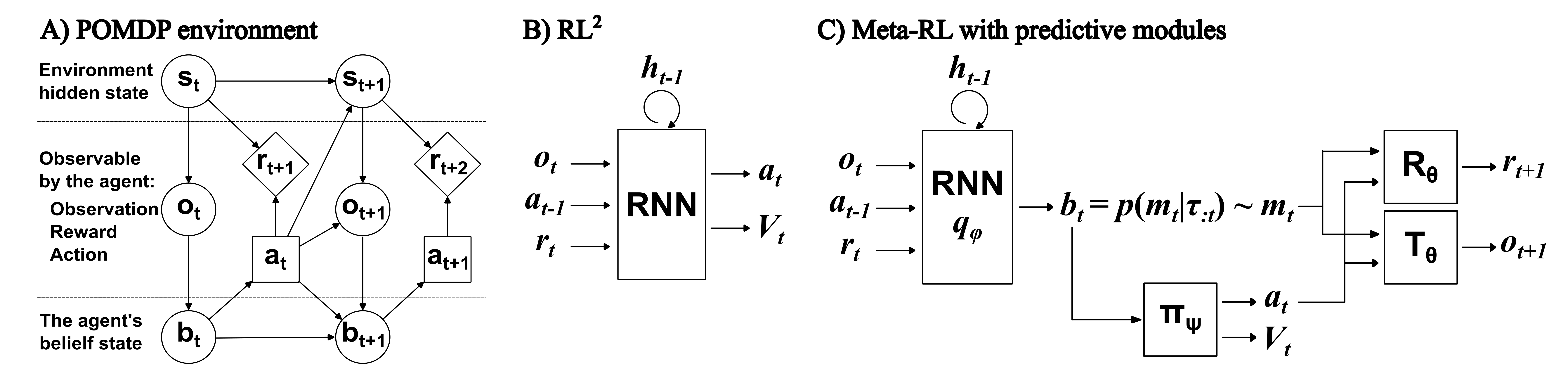}
  \caption{\textbf{Representation learning under partial observability using meta-RL with predictive modules.} 
  A) In POMDPs, a belief state \( b \), over the environment hidden state is maintained for policy learning. 
  B) Memory-based meta-RL (e.g. RL\textsuperscript{2}) simultaneously learns representation and policy using only the reward signal. 
  C) The proposed meta-RL with self-supervised predictive modules separates representation learning with predictive coding and policy learning with reward signal.}
  \label{fig:intro}
\end{figure}

\paragraph{Memory-based meta-RL}
Exemplified by RL\textsuperscript{2} \cite{duan2016rl, wang2016learning}, memory-based meta-RL emerges as a powerful method for learning adaptive policies across related tasks.
Often parameterized as recurrent neural networks (RNNs), RL\textsuperscript{2} implicitly maintains a memory of the past history in the recurrent states and learns a history-dependent policy by maximizing return over the task distribution using policy gradient algorithms (Fig. \ref{fig:intro}B).
This design enables the agent to adapt dynamically as it interacts with the environment, effectively learning an internal RL algorithm via its recurrent computation \cite{beck2023survey}.
Through end-to-end meta-learning over a task distribution, RL\textsuperscript{2} provides effective black-box agents that achieve rapid adaptation to new task instances and generalize across related tasks.

The training objective of meta-RL can be recast and understood as approximating the Bayes-optimal policy for a task distribution \cite{ortega2019meta}. 
Since history is sufficient for computing the belief state, black-box meta-RL can in principle learn Bayes-optimal policies by encoding the belief state in its hidden state. 
Several studies have demonstrated that belief states can be decoded from RNN hidden states \cite{akuzawa2021estimating, lambrechts2022recurrent, hennig2023emergence}.
Moreover, \citet{ortega2019meta} provided a theoretical framework for interpreting meta-RL as Bayes-optimal solutions, and \citet{mikulik2020meta} empirically showed that meta-RL agents learn near Bayes-optimal policy in bandit problems.
However, a comparison of meta-learned representations in RL\textsuperscript{2} using state machine simulation suggests that their representations are not equivalent to the minimally-sufficient Bayes-optimal states, likely due to failures in injectivity, which may hinder the agent's interpretability, robustness, and ability to generalize \cite{mikulik2020meta}.
In fact, in practice, black-box meta-RL may struggle to efficiently learn Bayes-optimal policies in tasks requiring exploration that is temporally-extended and qualitatively different from exploitation behavior \cite{zintgraf2021varibad, beck2023survey, liu2021decoupling}.

\paragraph{Predictive coding neural circuits}
In neuroscience, theoretical and experimental studies have explored predictive processing as a fundamental computational objective across brain regions and as a neural implementation of Bayesian computation. 
Sensory areas are often modeled as hierarchical neural networks predicting future sensory inputs \cite{rao1999predictive, furutachi2024cooperative, zhuang2021unsupervised}, potentially supporting efficient neural representation for perceptual inference \cite{ma2006bayesian}.
Motor systems are thought to predict the sensory consequences of the motor outputs \cite{kim2015cellular}, which can underlie efficient integration in sensorimotor learning \cite{kording2004bayesian}.
The hippocampus, with neural activity believed to forecast an animal’s upcoming experiences, is critical for relational learning and computing a predictive map or transition model to support efficient navigation \cite{schiller2015memory, whittington2020tolman}.
Building on the relationship between predictive learning and efficient neural representation, recent work has begun to explore adding predictive objectives to encourage artificial agents to develop activities similar to those observed in the brain \cite{fang2023predictive}. 

\paragraph{Predictive learning in meta-RL}
Recent work has explored alternative architectures or auxiliary training objectives to learn latent representations that are predictive of future observations or rewards for augmenting meta-RL agents.
For instance, \citet{igl2018deep} showed deep policy can be improved by adding predictive auxiliary loss, and proposed particle filter method for approximating belief inference. 
Several studies further decoupled learning recurrent generative models of the environment and used the learned latent representations as inputs for model-free RL algorithms, demonstrating improved performance especially for robotics tasks that are difficult for black-box models \cite{ha2018recurrent, han2019variational, gregor2019shaping, lee2020stochastic}.
For a special type of POMDP where hidden states are stationary, several works considered approximating task inference by taking advantage of privileged information \cite{humplik2019meta}, posterior sampling \cite{rakelly2019efficient}, or variational inference \cite{zintgraf2021varibad}, providing effective meta-RL models for rapid adaptation to new task instances.
Recent works have explored how to further improve representation learning through state-space modeling \cite{akuzawa2021estimating}, normalizing flow technique \cite{chen2022flow}, or decoupled belief modeling with separate random policies
\cite{wang2023learning}.
On the other hand, learning latent dynamical world models has also been proposed for model-based deep RL approaches, including SOLAR \cite{zhang2019solar}, PlaNet \cite{hafner2019learning} and Dreamer \cite{hafner2025mastering}. 
These studies demonstrated that learning latent world models through multi-step predictive objectives can strongly improve planning and performance in challenging control benchmarks.

Complementary to the above empirical progress of integrating predictive learning to improve performance, our work provides the interpretability foundation for why predictive objectives work in deep RL by focusing on understanding its meta-learned representation and computation.

\section{Meta-RL with self-supervised predictive modules}
\label{main_model_detail}

Black-box meta-RL like RL\textsuperscript{2} (Fig. \ref{fig:intro}B) requires simultaneously learning history representation and policy using only reward signals, potentially suffering from inadequate representation and policy learning \cite{beck2023survey, mikulik2020meta}.
To overcome this challenge, we employ an end-to-end meta-RL framework with self-supervised predictive modules to learn explicit belief representations. 
Our motivations are two-fold: algorithmically, a good representation should summarize the relevant history predictive of the future and reflect uncertainty; neurobiologically, modular neural circuits are believed to perform future predictions, attaining efficient representations under predictive coding \cite{rao1999predictive}.

The proposed model, meta-RL with predictive modules, consists of a variational autoencoder (VAE) for predictive representation learning and a policy network operating on the learned representation (Fig. \ref{fig:intro}C).
The predictive modules begin with an RNN encoder $q_\phi$, which takes as input current observation $o_t$, reward $r_{t}$, and previous action $a_{t-1}$.
The RNN outputs a low-dimensional bottleneck $b_t$ which is the posterior distribution over the latent states $m_t$ conditioned on the history $\tau_{:t}$.
The posterior is trained via reward and observation prediction akin to predictive coding using the reward decoder $R_{\theta}$ and the observation decoder $T_{\theta}$, respectively, to predict upcoming rewards and observations given the action taken by the policy network.
The predictive modules optimize the evidence lower bound (ELBO) using the reparameterization technique \cite{kingma2013auto}, with KL regularization loss similar to Bayesian filtering (see \ref{elbo_derivation} for details). 
This approach learns an explicit probabilistic belief representation $b_t$ over the latent state, facilitating a direct interpretation as belief states in POMDPs. 
The policy network $\pi_{\psi}$ is a feedforward neural network receiving the belief state $b_t$ as an input and can be efficiently trained using model-free policy gradient algorithms \cite{mnih2016asynchronous}. 
Note the policy gradient from RL loss is only used to train the policy network but not the RNN encoder.
The entire model is trained in a self-supervised, end-to-end manner using the standard meta-learning paradigm. 

Our parameterization is similar in principle to previous work \cite{zintgraf2021varibad, akuzawa2021estimating, han2019variational}, with designs tailored for POMDPs without strict assumptions over stationarity \cite{zintgraf2021varibad} and structures \cite{akuzawa2021estimating} of the hidden states, facilitating a direct interpretation of $b_t$ as the belief states in POMDPs.
Furthermore, the end-to-end approach neither relies on privileged information \cite{humplik2019meta}, nor requires a separate random policy to generate samples for training predictive models \cite{wang2023learning} as exploited in previous work.

\section{Experiment}
\label{experiment_methods}

We designed the following experiments to answer the central question: Does meta-RL with self-supervised predictive modules enhance learning representations that more closely match the Bayes-optimal belief states than black-box meta-RL? 
Specifically, we adopt the \emph{state machine simulation} analysis used in \citet{mikulik2020meta} to examine the equivalence of representation and computation between meta-RL agents and Bayes-optimal solutions.
In essence, two state machines can be considered computationally equivalent if they can \emph{simulate} each other---that is, if for any given state in one machine, we can find a mapping onto the other machine such that both their \emph{state transitions} and \emph{outputs} are the same \cite{girard2005approximate}.
This analysis examines whether there exists a consistent way of interpreting every state in one machine as a state in the other. 
Although decoding is commonly used to evaluate representation similarity \cite{akuzawa2021estimating, lambrechts2022recurrent, hennig2023emergence}, it only measures correlations between representations. 
In contrast, state machine simulation thoroughly assesses representational equivalence based on structural and computational relevance, providing a rigorous analysis for model interpretability.

\paragraph{State machine simulation} 
In tasks where Bayes-optimal states are computationally tractable, we can compare meta-RL representations with Bayes-optimal belief states using state machine simulation \cite{mikulik2020meta}.
The goal is to measure the \emph{state} and \emph{output dissimilarities} after mapping the states from one machine to the other.
If both state and output dissimilarities are low in \emph{both} mapping directions, then the two representations can be considered computationally equivalent.
Briefly, the procedures are as follows (and see \ref{sma} for details): (i) two mapping functions (parameterized as multi-layered perceptrons) are trained to map from meta-learned states to Bayes-optimal states and from Bayes-optimal states to meta-learned states; (ii) The \emph{state dissimilarity} $D_s$ is measured as the mean square error (MSE) of the mapped states against the target states; (iii) The \emph{output dissimilarity} $D_o$ is measured as the difference in return as generated by the output of the original states and the output of the mapped states. 
Here we compare $D_s$ and $D_o$ of meta-RL with predictive modules against those of black-box meta-RL (RL\textsuperscript{2}) to evaluate the quality of their learned representations.

\begin{figure}
  \centering
  \includegraphics[width=1.0\textwidth]{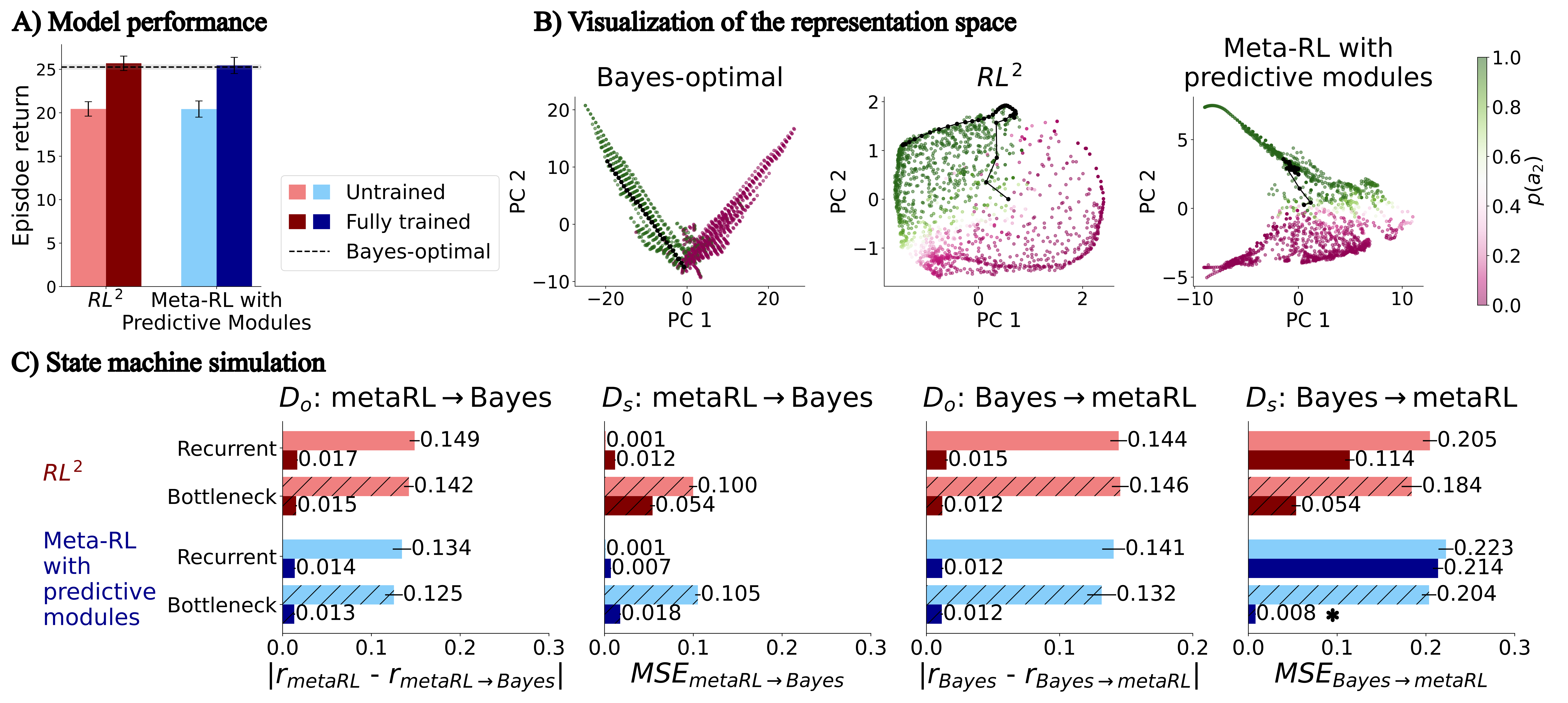}
  \caption{\textbf{Meta-RL with predictive modules better approximates Bayes-optimal states in bandit task.} 
  A) Model performance against the Bayes-optimal policy. 
  Light and dark colors denote untrained and fully trained models, respectively. 
  B) Visualization of Bayes-optimal belief states, learned representation in the bottleneck layer of RL\textsuperscript{2}, and that of meta-RL with predictive modules. 
  States are colored by the probability of choosing $a_2$. 
  Black curves show one example trajectory. 
  C) State machine simulation.  
  \emph{Recurrent} and \emph{bottleneck} denote the layers used for analysis.
  The bottleneck layer of meta-RL with predictive modules achieves the lowest state dissimilarity $D_s$ and output dissimilarity $D_o$ in both mapping directions, indicating highest equivalence to Bayes-optimal states.
  (For all figures error bars denote the s.e.m. across at least 5 training runs, and asterisks denote statistical significance ($p<0.05$, t-test) among the trained models.) }
  \label{fig:stationary_bandit}
\end{figure}

\paragraph{Tasks} 
To enable rigorous comparison between meta-RL agents and Bayes-optimal solutions, we select the following exemplar tasks for which Bayes-optimal solutions are computationally tractable. 
While previous work has only evaluated stationary bandits \cite{mikulik2020meta}, our task suite covers diverse POMDP challenges: explore-exploit tradeoff, sequential decision-making, dynamic belief tracking, information gathering, and continuous control. 
To our knowledge, this is the first systematic evaluation of meta-RL representations using ground-truth comparison across diverse POMDPs.
Following are the six classes of tasks evaluated (details of each task are described in \ref{task_details}):

\begin{itemize}
    \item \textbf{Classic two-armed Bernoulli bandit}: 
    The first task is the classic multi-armed bandit also considered in \citet{mikulik2020meta}, where the meta-learned representations in RL\textsuperscript{2} models were found to deviate from the minimally sufficient Bayes-optimal states even though their policy converge to the Bayes-optimal one. 
    We will examine if the proposed meta-RL with predictive modules can effectively learn Bayes-optimal belief states in two-armed bandits.

    \item \textbf{Dynamic two-armed bandit}: 
    To evaluate dynamic belief tracking, we next consider a dynamic extension of the two-armed bandit.
    Each arm is in one of $n$ possible discrete states (for simplicity we choose $n=2$) with various reward probabilities.
    The state of each arm evolves according to an independent Markovian transition process.
    The task is designed such that the belief state update is analytically tractable, and that the Bayes-optimal policy can be derived using the value iteration algorithm over the belief space \cite{Sutton1998}.
    To examine different structural and dynamical configurations, we consider three parameter settings: (i) two arms of symmetric reward probability states and transition dynamics, (ii) two arms of asymmetric reward probability states, and (iii) two arms of asymmetric transition dynamics.
    
    \item \textbf{Stationary Tiger}: 
    To exemplify sequential decision-making under uncertainty, we consider the classic POMDP Tiger task \cite{kaelbling1998planning}, where an agent chooses between two doors---one hiding a tiger (penalty=\(-100\)) and the other hiding a treasure (reward=\(10\)). 
    The agent may additionally choose to pay a small penalty=\(-1\) to ``listen'' to acquire noisy observations about the tiger's location. 
    This simple yet powerful paradigm tests an agent's ability to balance information gathering with reward seeking.
    We consider two difficulty levels by varying the observation accuracy for the ``listen'' action.

    \item \textbf{Dynamic Tiger}:
    We extend the Tiger task to a dynamic version by allowing the tiger's location to change over time, following Markov transition dynamics. 
    As information reliability and relevance are further corrupted by the dynamic nature of the hidden state, the agent has to balance listening more to increase confidence against making decisions earlier in case the information gathered becomes obsolete. 
    This task design permits tractable belief updates and the Bayes-optimal policy can be derived using value iteration \cite{Sutton1998}. 
    Similarly, we consider two difficulty levels by changing the observation accuracy.
    
    \item \textbf{Oracle bandit}: 
    Black-box meta-RL often struggles when exploration and information seeking are required \cite{beck2023survey}. 
    We hypothesize this is due to ineffective representation learning. 
    To exemplify this point, we consider an illustrative \emph{oracle bandit}: in an 11-arm bandit environment, one of the first ten arms $a_{1-10}$ is the target arm that gives a payout of 5, whereas the remaining nine are non-target arms with a payout of 1. 
    The last arm $a_{11}$ is the ``oracle'' arm whose payout is $\leq 1$ but informs the target arm in the form of $\frac{1}{10}$ of the target arm index (e.g. a reward of 0.3 from $a_{\text{11}}$ indicates $a_3$ is the target arm). 
    This is similar to tasks discussed in \citet{wang2016learning} but differs in that here no structured feedback is given, which makes representation learning more challenging and critical. 
    A successful policy requires paying an immediate exploration cost to acquire information for long-term gain.

    \item \textbf{Latent goal cart}: 
    Deriving Bayes-optimal solutions in continuous POMDPs are usually intractable, hindering rigorous representational equivalence analysis.
    To enable evaluation in continuous observation and action spaces, we design an exemplar continuous control task which still permits tractable Bayes-optimal belief inference and policy.
    In this task, an agent controls a continuous action, the velocity of a cart, to move along a 1-dimensional track to a hidden goal (\(+1\) or \(-1\)) which needs to be inferred from the continuous observation (current position) and reward (noisified distance from the hidden goal) it receives.

\end{itemize}

\paragraph{Agents}
For each POMDP environment, two types of agents---the proposed meta-RL with predictive modules and the black-box meta-RL (i.e. RL\textsuperscript{2})---were trained on the target task distribution.
To facilitate comparison, the baseline RL\textsuperscript{2} model is designed to be architecturally identical to meta-RL with predictive modules except for the decoder networks.
Specifically, the RL\textsuperscript{2} baseline consists of an RNN followed by a fully connected bottleneck layer which is the counterpart to the latent belief layer $b_t$ in Fig.\ref{fig:intro}C, and finally with a readout layer that generates action logits and value estimates (details see \ref{agent_details}).
Identical layer sizes are used across both models throughout each experiment. 

While the latent belief layer (i.e. the bottleneck layer) for meta-RL with predictive modules is the natural target to be interpreted as belief states, \citet{mikulik2020meta} evaluates the representations learned in the recurrent layers in RL\textsuperscript{2} models.
In what follows, we systematically analyze the representations learned in both the \emph{recurrent} and \emph{bottleneck} layers in each model using state machine simulation.

\section{Results}
\label{results}

\paragraph{Two-armed Bernoulli bandit} 
The Bayes-optimal solution can be derived using the Gittins index method \cite{gittins1979bandit}. 
After training, both RL\textsuperscript{2} and meta-RL with predictive modules approach Bayes-optimal return (Fig. \ref{fig:stationary_bandit}A).
Visualization of the bottleneck layers shows meta-RL with predictive modules learns a low-dimensional representation structurally more similar to the Bayes-optimal states than RL\textsuperscript{2} (Fig. \ref{fig:stationary_bandit}B). 
This observation is corroborated by the state machine simulation analysis results (Fig. \ref{fig:stationary_bandit}C). 
Before training, both models have high $D_s$ and $D_o$, indicating the untrained representations are far from Bayes-optimal. 
One notable exception is $D_s$ of the recurrent layers for the meta-RL$\rightarrow$Bayes mapping direction, which further verifies that using decoding alone is not enough for evaluating the equivalence of two representations, as the untrained recurrent layers may maintain a verbose representation of the history and can be mapped arbitrarily close to the belief states if powerful enough mapping functions are used (see \ref{sma} for detailed discussions).

After training, for the recurrent layers in both models, the state dissimilarity $D_s$ remains high for the Bayes$\rightarrow$meta-RL direction, highlighting the representations in the recurrent layer deviate from the minimally-sufficient Bayes-optimal belief states, similar to the findings in \citet{mikulik2020meta}.
In contrast, considering all four measures, the bottleneck layer in meta-RL with predictive modules achieves the lowest $D_s$ and $D_o$ in both mapping directions, indicating that its learned representations best approximate the Bayes-optimal belief states (Fig. \ref{fig:stationary_bandit}C).
Remarkably, the representation learning capacity does not merely arise from having a low-dimensional bottleneck, as $D_s$ for the bottleneck layer in RL\textsuperscript{2} remains significantly higher than that in meta-RL with predictive modules, suggesting that predictive coding encourages more compact, efficient representations in meta-RL agents.

\paragraph{Dynamic bandits} 
\begin{figure}
  \centering
  \includegraphics[width=1.0\textwidth]{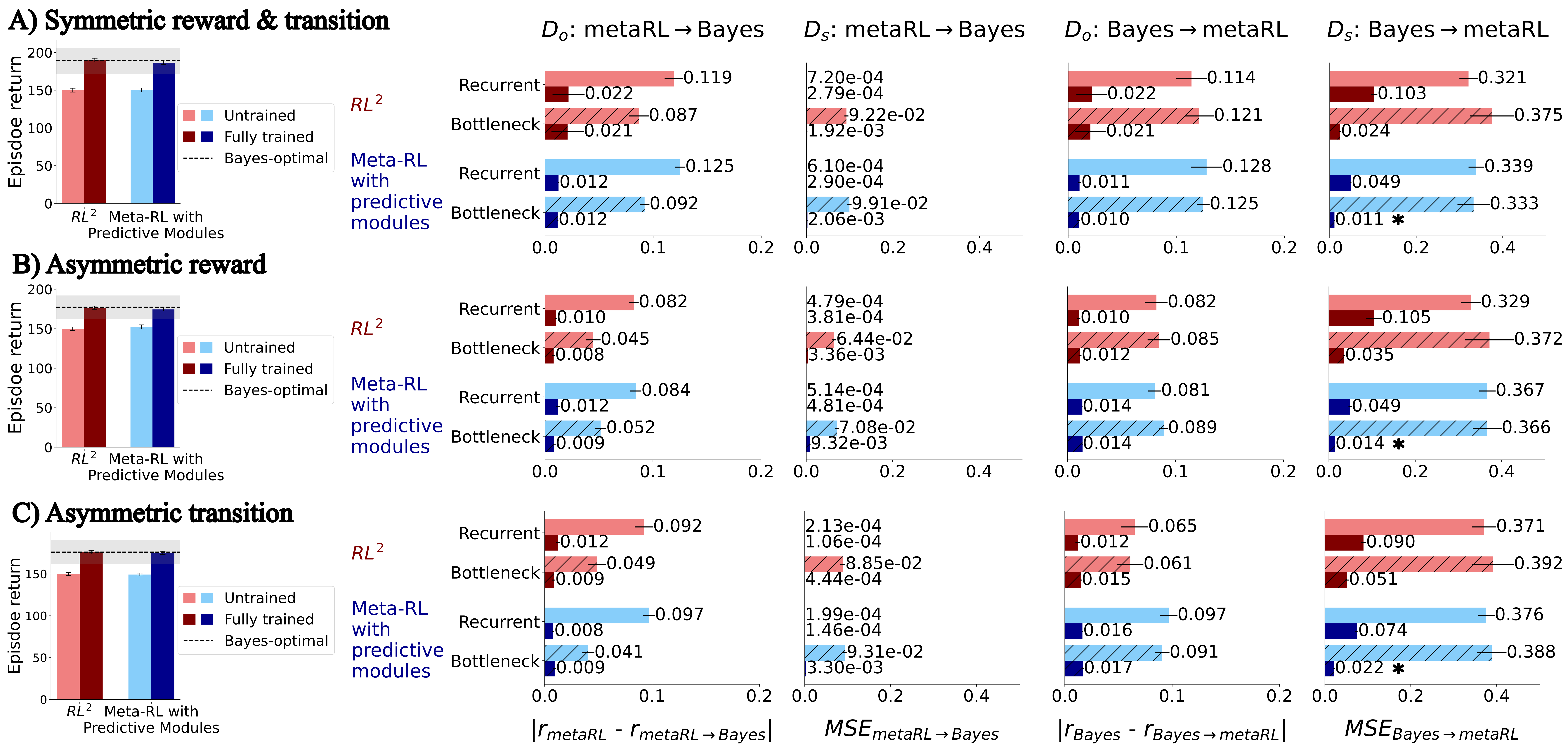}
  \caption{\textbf{Dynamic bandits.} 
  Performance (left) and state machine simulation analysis (right) for A) the symmetric task, B) the asymmetric reward task, and C) the asymmetric transition task.
  Shaded areas in the left column denote the standard deviation of episodic return under Bayes-optimal policy.
  The bottleneck layer of meta-RL with predictive modules attains the lowest state dissimilarity $D_s$ and output dissimilarity $D_o$ in both mapping directions across tasks.
  }
  \label{fig:dynamic_bandit}
\end{figure}

Across the three dynamic bandit tasks---symmetric reward and transition, asymmetric reward, and asymmetric transition (Fig. \ref{fig:dynamic_bandit}A, B, and, C, respectively), both models approach the optimal return.  
Meta-RL with predictive modules learns representations structurally more similar to the Bayes-optimal states (see \ref{append_state_space_visualization}).
State machine simulation reveals that the recurrent layers in both models present with large $D_s$ for the Bayes$\rightarrow$meta-RL direction even after training. 
While the bottleneck layers in both models achieve comparably low output dissimilarities, meta-RL with predictive modules consistently achieve significantly lower state dissimilarity $D_s$ in the Bayes$\rightarrow$meta-RL mapping direction, indicating a higher equivalence to the Bayes-optimal belief states.

\paragraph{Stationary Tiger and Dynamic Tiger} 
\begin{figure}
  \centering
  \includegraphics[width=1.0\textwidth]{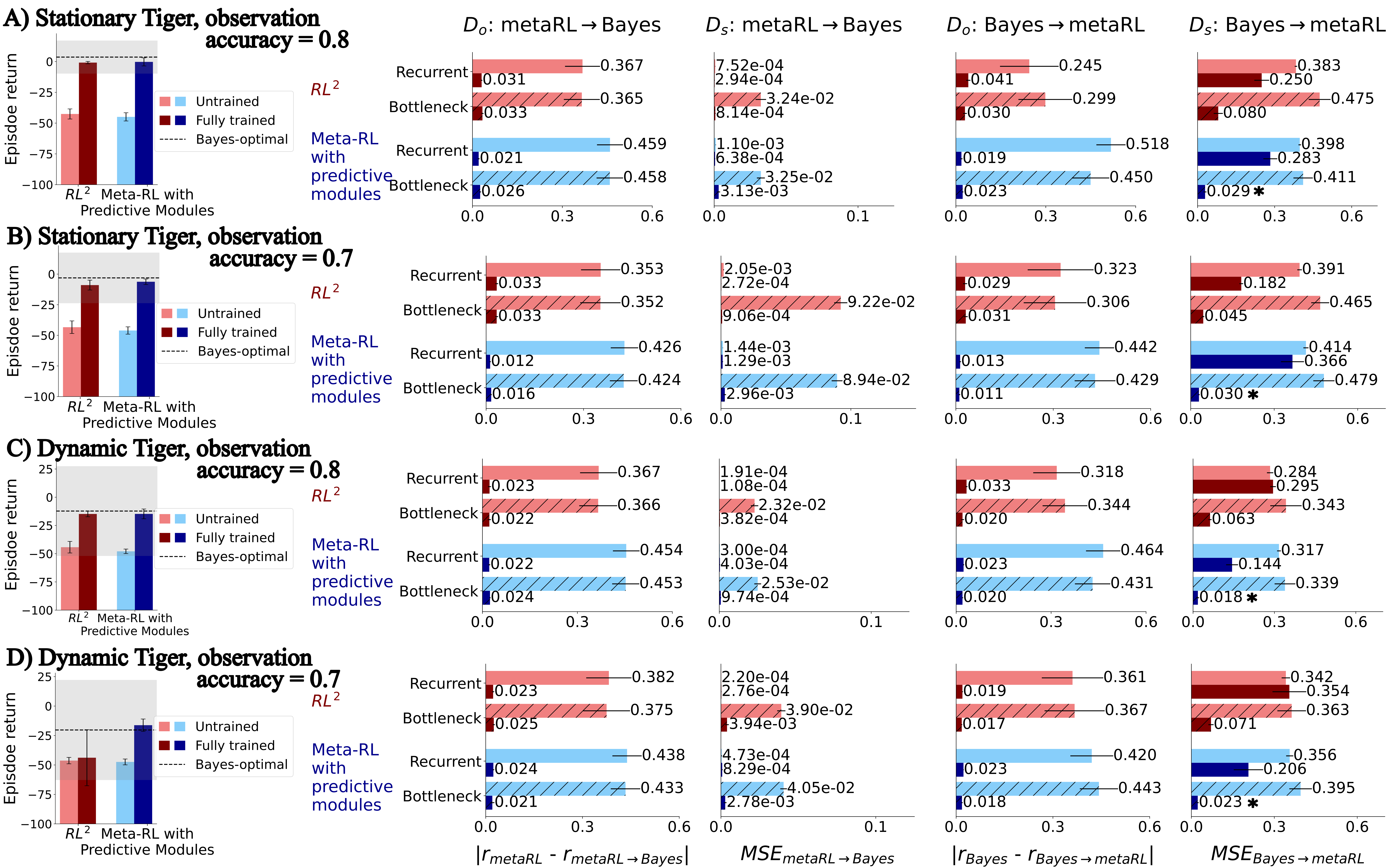}
  \caption{\textbf{Stationary and dynamic Tiger.} 
  Performance (left) and state machine simulation analysis (right) for A--B) stationary Tiger with two levels of observation accuracies, and C--D) dynamic Tiger, where the tiger location changes with a probability of $0.9$ at each timestep.  
  Note for the most challenging setting in D, only meta-RL with predictive modules approaches the Bayes-optimal return.
  }
  \label{fig:tiger_result}
\end{figure}

Difficulty of the Tiger tasks increases when the observation accuracy decreases (information becomes less reliable) and when the tiger's location is dynamic (information may become obsolete if waiting for too long).
After training, both models approach the Bayes-optimal return in easier tasks, while for the most challenging dynamic Tiger with low observation accuracy, only meta-RL with predictive modules approach Bayes-optimal return (Fig. \ref{fig:tiger_result}D).
Meta-RL with predictive modules learns representations better capturing the task structure (see \ref{append_state_space_visualization}).
State machine simulation shows that the recurrent layers in both models fail to approximate Bayes-optimal belief states (large $D_s$ after training in Fig. \ref{fig:tiger_result}).
Comparing the bottleneck layers, meta-RL with predictive modules consistently achieves higher equivalence to Bayes-optimal belief states, as indicated by the significantly lower $D_s$ for the Bayes$\rightarrow$meta-RL direction across all tasks.

\begin{figure}[h!]
  \centering
  \includegraphics[width=1.0\textwidth]{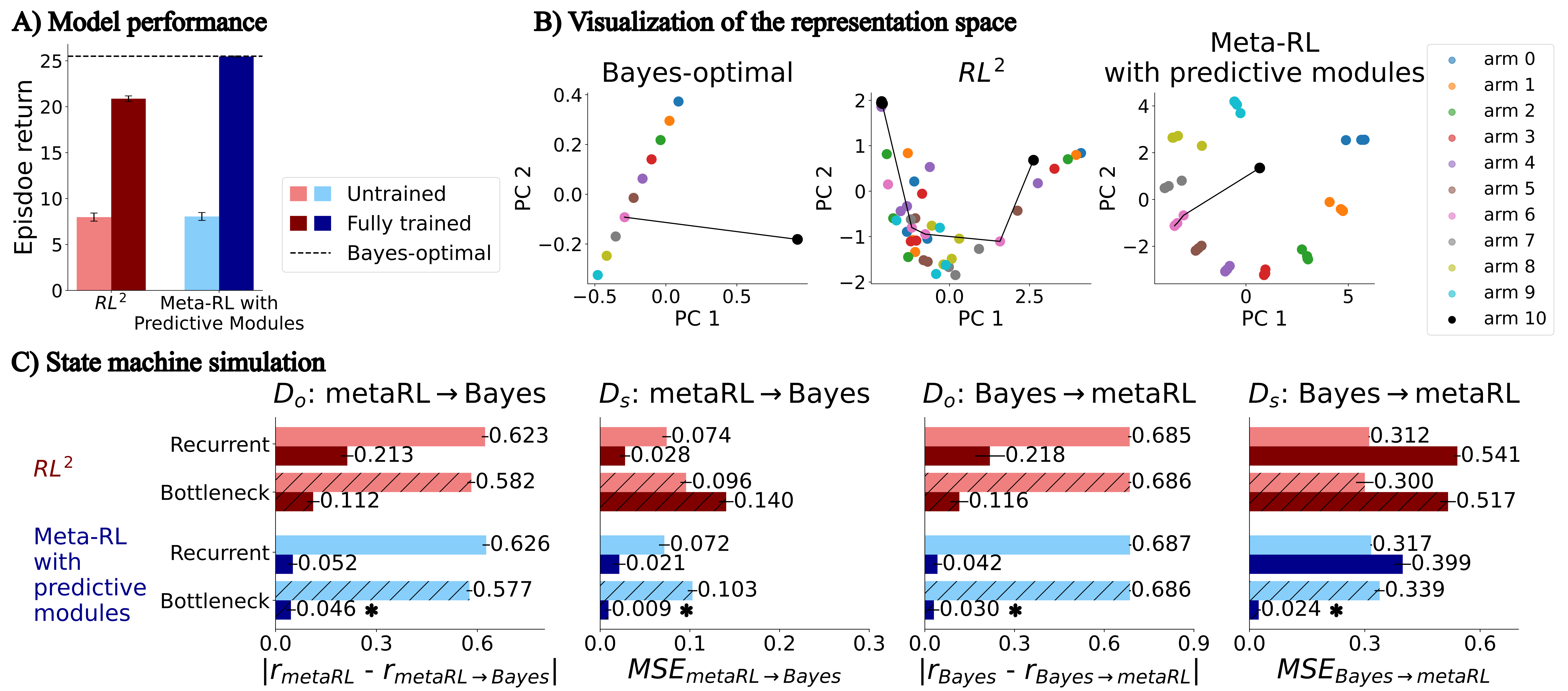}
  \caption{\textbf{Meta-RL with predictive modules learns interpretable Bayes-optimal solutions in oracle bandit.} 
  A) Only meta-RL with predictive modules learns the optimal policy. 
  B) Bayes-optimal states (left), the bottleneck layer of RL\textsuperscript{2} (middle), and that of meta-RL with predictive modules (right). 
  States are colored by the most likely arm to choose. 
  C) The bottleneck layer in meta-RL with predictive modules attains the lowest state and output dissimilarities in both mapping directions. 
  }
  \label{fig:oracle_bandit_result}
\end{figure}

\paragraph{Oracle bandit}
After training, only meta-RL with predictive modules learns the Bayes-optimal policy (Fig. \ref{fig:oracle_bandit_result}A), paying an immediate cost to sample the oracle arm and utilize the information for long-term gain.
In contrast, RL\textsuperscript{2} converges to a suboptimal policy where it learns to sample the oracle arm upfront, but fails to use the attained information consistently.
Visually, the bottleneck layer in meta-RL with predictive modules shows interpretable representations capturing task structures and dynamics. 
In contrast, RL\textsuperscript{2} fails to learn an interpretable representation (Fig. \ref{fig:oracle_bandit_result}B), likely contributing to its suboptimal behavior. 
State machine simulation shows that after training, the bottleneck layer in meta-RL with predictive modules achieves significantly lower state and output dissimilarities in both mapping directions, indicating close approximation of Bayes-optimal belief states (Fig. \ref{fig:oracle_bandit_result}C).
Dissimilarity measures remain high for RL\textsuperscript{2} after training, suggesting that RL\textsuperscript{2} may suffer from inadequate representation learning when the task requires exploration and information seeking.

\paragraph{Latent goal cart}

\begin{figure}[h]
  \centering
  \includegraphics[width=1.0\textwidth]{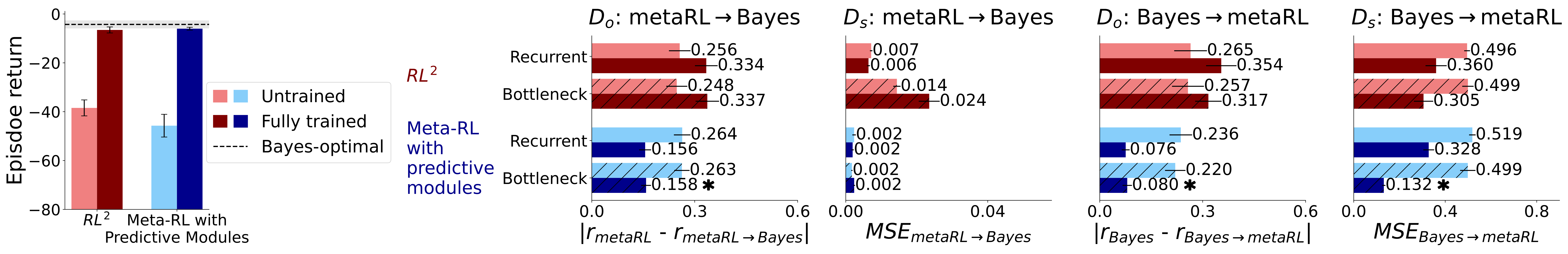}
  \caption{\textbf{Latent goal cart.} 
  Performance (left) and state machine simulation analysis (right).
  The bottleneck layer of meta-RL with predictive modules attains the lowest state and output dissimilarities in both mapping directions, indicating close approximation of Bayes-optimal belief states. 
  }
  \label{fig:latent_goal_cart}
\end{figure}

After training, both models approach the Bayes-optimal return (Fig. \ref{fig:latent_goal_cart}).
Visualization of the states shows meta-RL with predictive modules learns representations better capturing the task structure (see \ref{append_state_space_visualization}).
State machine simulation shows that the recurrent layers in both models fail to approximate Bayes-optimal belief states (large $D_s$ after training).
In contrast, the bottleneck layers of meta-RL with predictive modules achieve the highest equivalence to Bayes-optimal belief states, as indicated by the significantly lower state and output dissimilarities in both mapping directions.

\begin{figure}[h!]
  \centering
  \includegraphics[width=1.0\textwidth]{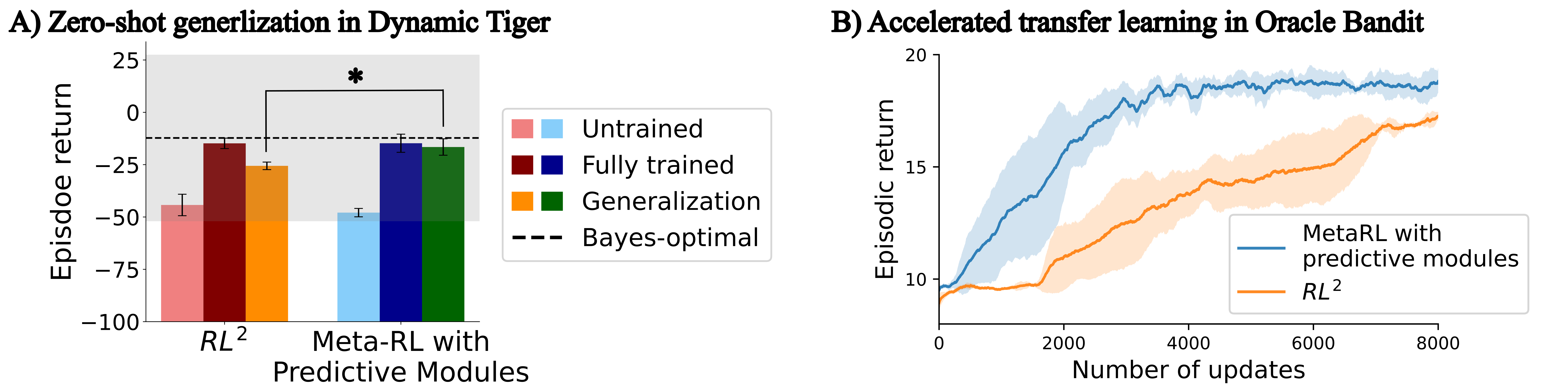}
  \caption{\textbf{Better representation leads to improved generalization capability.} 
  A) When testing models trained on dynamic Tiger with observation accuracy of 0.7 on environments with observation accuracy of 0.8, meta-RL with predictive modules shows near-optimal zero-shot test return (\(-15.94\pm3.82\), green), whereas RL\textsuperscript{2} has significantly lower test return (\(-25.56\pm1.80\), orange), indicating that meta-RL with predictive modules achieves better zero-shot generalization to unseen but relevant tasks.
  (Red and blue bars are reference model performance from Fig. \ref{fig:tiger_result}C.)
  B) When models are first trained on oracle bandits with a task distribution that only contain arms 1-5 before transferring to a second task distribution that contains arms 6-10, meta-RL with predictive modules demonstrates significantly faster transfer learning as compared to RL\textsuperscript{2}, indicating that meta-RL with predictive modules achieves more efficient transfer learning to out-of-distribution relevant tasks.
  }
  \label{fig:generalization}
\end{figure}

\paragraph{Improved generalization capability}
Thus far we present a systematic evaluation of representational equivalence between meta-RL with predictive modules and Bayes-optimal solutions across a diverse array of POMDP tasks.
Our state machine simulation results strongly suggest that meta-RL with predictive modules can effectively learn interpretable representations that better approximate Bayes-optimal belief states, which in turn could lead to improved policy learning in challenging tasks.
To further illustrate the importance and implications of learning Bayes-optimal belief state representations in partially observable environments, we examine whether learning better representations might lead to improved generalization capability to unseen but related tasks.
To this end, we first evaluate zero-shot generalization capacity.
As shown in Fig. \ref{fig:generalization}A, when testing models trained on dynamic Tiger with observation accuracy of 0.7 on environments with observation accuracy of 0.8, meta-RL with predictive modules shows near-optimal zero-shot test return, whereas RL\textsuperscript{2} has significantly lower test return, suggesting that through learning better representations, meta-RL with predictive modules could capture the underlying belief updates shared across similar tasks to facilitate zero-shot generalization.
Next, we evaluate whether representation quality impacts transfer learning efficiency in out-of-distribution (OOD) tasks.
As shown in in Fig.\ref{fig:generalization}B, when models are first trained on oracle bandits with a pre-training distribution that only contain arms 1-5 before transferring to a second task distribution that contains arms 6-10, meta-RL with predictive modules shows significantly faster transfer learning as compared to RL\textsuperscript{2}.
This implies that with better representation learning, meta-RL with predictive modules could effectively learn the relevant environmental structure and dynamics to facilitate more efficient transfer learning to OOD tasks.

\paragraph{Ablation studies}
Finally, to pinpoint which algorithmic design choices drive the enhanced representation learning capacity in meta-RL with predictive modules, we performed two targeted ablations: (i) no KL: we remove the VAE’s KL regularization to test whether latent regularization is necessary to enforce compact representations, and (ii) joint RL: we allow the policy gradients from RL loss to jointly train the RNN encoder alongside the predictive loss---opposed to training the encoder \emph{solely} with the predictive loss as our proposed model does---to assess whether policy gradients provide additional representational benefits.
Table \ref{ablation study} summarizes changes in dissimilarities from state machine simulation analysis on exemplar tasks (see \ref{append_sms_full_results} and \ref{append_sensitivity_analysis} for full results).
First, without KL regularization on the latent space, bottleneck dissimilarities to Bayes‑optimal states increase (i.e. worse alignment) on most tasks (except for the minor decrease in dynamic bandits), showing that \emph{suitable} regularization may help to promote compact, belief‐like representations.
On the other hand, under joint RL training, bottleneck dissimilarities are not further decreased as compared to the predictive loss‐only models, indicating that additional policy gradients do not further improve representations and hence are not necessary for learning Bayes-optimal representations.
This indicates that the enhanced representation learning capacity can be largely attributed to predictive learning.

\begin{table}
  \caption{Ablation studies: 
  changes in bidirectional state and output dissimilarities of the bottleneck layers compared to the proposed model of meta-RL with predictive modules. 
  ($M$ for meta-RL, $B$ for Bayes-optimal solutions. 
  Bold indicates $p<0.05$.)
  }
  \label{ablation study}
  \footnotesize
  \setlength{\tabcolsep}{2.5pt}
  \centering
  \begin{tabular}{lcccccccc}
    & $\Delta D_o^{M \rightarrow B}$ & $\Delta D_s^{M \rightarrow B}$ & $\Delta D_o^{B \rightarrow M}$ & $\Delta D_s^{B \rightarrow M}$ & $\Delta D_o^{M \rightarrow B}$ & $\Delta D_s^{M \rightarrow B}$ & $\Delta D_o^{B \rightarrow M}$ & $\Delta D_s^{B \rightarrow M}$ \\
    \cmidrule(l){2-5} \cmidrule(l){6-9}
    
    & \multicolumn{4}{c}{Two-arm Bernoulli bandit (unit: 1e-2)} & \multicolumn{4}{c}{Dynamic bandit (unit: 1e-2)} \\
    No KL & 
    0.12±0.4 & \textbf{2.44±0.38} & -0.26±0.36 & \textbf{0.21±0.04} &
    -0.16±0.53 & 0.42±0.38 & -0.13±0.38 & \textbf{-0.65±0.15}  \\
    joint RL & 
    \textbf{-0.41±0.05} & -0.61±0.15 & -0.27±0.12 & -0.08±0.04 &
    -0.09±0.06 & 0.08±0.13 & 0.0±0.04 & 0.47±0.2  \\
    
    \midrule

    & \multicolumn{4}{c}{Dynamic Tiger (unit: 1e-2)} & \multicolumn{4}{c}{Oracle bandit (unit: 1e-2)} \\
    No KL &
    1.11±1.35 & 0.22±0.17 & \textbf{1.49±0.6} & 1.11±1.76  &
    \textbf{9.22±4.44} & \textbf{23.47±6.94} & \textbf{56.3±8.43} & 0.24±0.82  \\
    joint RL &
    1.1±0.96 & \textbf{0.21±0.05} & 0.95±1.66 & 0.62±0.84  &
    \textbf{10.91±1.02} & 0.24±0.82 & 5.53±5.0 & 1.77±3.38  \\
    
    \bottomrule
  \end{tabular}
\end{table}

\section{Discussion and conclusions}
\label{discussion_and_conclusions}
In this work we employed rigorous state machine simulation analysis to demonstrate that meta-RL with self-supervised predictive coding modules can effectively learn interpretable, near Bayes-optimal belief state representations across diverse partially observable tasks, whereas conventional black-box meta‑RL fails to capture the minimally sufficient representation. 
We further showed how the enhanced representation learning capacity in meta-RL with predictive modules not only supports effective information gathering and policy learning in challenging tasks, but also leads to improved generalization capability to unseen but relevant tasks. 
Our results align with predictive coding theories in neuroscience, suggesting that predictive objectives constitute a general computational strategy for achieving Bayesian computation, and highlight predictive learning as a fundamental principle for guiding efficient representation learning in agents navigating partially observable environments.

\paragraph{Scope and limitations} 
We considered POMDPs with tractable Bayes-optimal solutions to enable rigorous analysis. 
Future work on more scalable analysis methods for tasks with larger state space will be valuable. 
In addition, we examined next-step future prediction as a general predictive objective. 
Considering the empirical success of multi-step predictive objective in deep RL, one promising direction is to investigate how incorporating multi-step, different time scales, and temporal abstractions into predictive modules might further enhance representation learning. 
Finally, we showed representation learning affects an agent's ability for generalization, but a more comprehensive evaluation is needed to explore how to meta-learn representations to support OOD generalization. 

\paragraph{Broader Impact}
Our work advances model interpretability by linking meta‑learned representations to Bayes‑optimal beliefs, offering a Bayesian account for the success of predictive objectives in deep RL. 
This understanding is critical for developing safe, adaptable agents that operate under uncertainty and limited information---such as in medical decision support and disaster response.
Moreover, our approach shows how insights from neuroscience theories may contribute to opening the black box of modern AI.
Conversely, uncovering the computational principles that drive representation learning in artificial agents may in turn shed light on the mechanisms of neural computation in the brain.

\newpage

\paragraph{Acknowledgements}
We thank Patrick Zhang and Matthew Storm Bull for
their insightful feedback and discussions.
We gratefully acknowledge the community support from the University of Washington Computational Neuroscience Center and the Allen Institute for Neural Dynamics.
We also thank the anonymous reviewers for their valuable and constructive review. 

\paragraph{Code availability}
The codebase for model training and state machine simulation is available at: \href{https://github.com/walkerlab/metaRL-predictive-representation}{https://github.com/walkerlab/metaRL-predictive-representation}

\bibliographystyle{unsrtnat}
\bibliography{neurips_2025}

\newpage
\appendix

\section{Technical Appendices and Supplementary Material}

\subsection{Evidence lower bound (ELBO)}
\label{elbo_derivation}
At a given time step $t$, the learning objective of the RNN encoder $q_{\phi}$ is to maximize
\begin{equation} \label{evidence}
    \mathbb{E}_{\rho(\tau_{0:t})}[ \log p_{\theta}(o_{t+1}, r_{t+1} | a_{0:t}) ]
\end{equation}

where $\rho(\tau_{0:t})$ is the distribution of trajectories generated by the policy $\pi_{\psi}$. 
Given that Eq. \ref{evidence} is intractable, we can instead optimize a tractable evidence lower bound (ELBO), computed as:

\begin{equation} 
\label{formula:elbo}
    \begin{split}
        \mathcal{L} = \mathbb{E}_{\rho(\tau_{0:t})} \Biggl[ \mathbb{E}_{ \Pi_{t=0}^{T-1} q_{\phi}(m_t|\tau_{0:t}) } \sum_{t=0}^{T-1} \Bigl\{ & [\log p_{\theta}(o_{t+1} | m_t, a_t) + \log p_{\theta}(r_{t+1} | m_t, a_t)] \\
        & - D_{KL}[q_{\phi}(m_t|\tau_{0:t}) || p_{\theta}(m_t) ] \Bigr\} \Biggr]
    \end{split}
\end{equation}

The terms $\mathbb{E}_q [\log p_{\theta}(o_{t+1} | m_t, a_t) + \log p_{\theta}(r_{t+1} | m_t, a_t)]$ are the future predictive loss, and the term $D_{KL}[q_{\phi}(m_t|\tau_{0:t}) || p_{\theta}(m_t)]$ is a regularization term using Kullback-Leibler (KL) divergence between the variational posterior $q_\phi(m_t|\tau_{0:t})$ and the prior over the latent variables $p_{\theta}(m_t)$. 
To encourage the variational autoencoder(VAE) to approximate Bayesian filtering for belief update, the prior is set to the previous posterior $q_\phi(m_{t-1} | \tau_{0:t-1})$ with the initial prior $q_\phi(m_0 | \tau_0)= \mathcal{N} (0, I)$. 
To optimize the ELBO (Eq. \ref{formula:elbo}), the expectation is approximated with Monte Carlo sampling as in standard VAE training \cite{kingma2013auto}.

\subsection{Task details}
\label{task_details}

\paragraph{Two-armed Bernoulli bandit} 
We consider a two-armed Beta-Bernoulli bandit task with an episode length of 40 time steps. 
At the beginning of each episode, $\theta_1$ and $\theta_2$, the reward probability biases for the two arms, are independently drawn from a fixed Beta distribution, $p(\theta_a) = Beta(1,1)$. 
$\theta_1$ and $\theta_2$ are then used to define the Bernoulli reward distributions for arm $a \in \{1,2\}$, respectively. 
The reward biases $\theta_a$ are hidden from the agent. 
At each time step, the agent chooses an arm $a \sim \pi$ sampled from its policy, and receives a binary reward sampled from $Ber(\theta_a)$, i.e. $r_t \sim p(r| \theta_a) = Ber(\theta_a)$. 
The discounted cumulative return is computed with a discount factor $\gamma=0.95$.
For multi-armed bandit tasks, Bayes-optimal policy can be derived using the Gittins index method \cite{gittins1979bandit}, and the minimally sufficient Bayes-optimal belief state is to keep track of the count of total pulls and the count of rewarded pulls of each arm.
In other words, for the two-armed bandit task, the Bayes-optimal belief states are 4-dimensional: $(n_{a_1}, n_{r_{a1}}, n_{a_2}, n_{r_{a2}})$, where $n_a$ and $n_{r_a}$ denote the count of total pulls and the count of rewarded pulls for arm $a$, respectively. 

\paragraph{Dynamic two-armed bandit}
To evaluate the meta-learned representation and behavior when the underlying environmental state can change over time, we consider a dynamic version of the two-armed bandit.
At a given time step, each arm is in one of $n$ possible discrete states.
For simplicity we choose $n=2$.
We denote the hidden state of arm $a$ at time $t$ as $s^a_t$, where $s^a_t \in \{\theta^a_1, \theta^a_2\}, \forall a \in \{1,2\} $, with each state associated with a different reward probability, i.e., $r_t^a \sim p(r| s^a_t) = Ber(s^a_t)$.
The state of each arm evolves according to an independent Markovian transition process independent of the action chosen: $p(s^a_{t+1} | s^a_{t}, a^{'}) = p(s^a_{t+1} | s^a_{t}) = T^a_{s^a_{t}, s^a_{t+1}} $.
The episode length is set to be $300$ time steps. 
To examine different structural and dynamical configurations, we consider three parameter settings: 
\begin{itemize}
    \item Symmetric reward and transition: the two arms share the same set of discrete states where $s^a_t \in \{0.1, 0.9\}, \forall a \in \{1,2\}$, and the same transition dynamics where $T^a_{0.1, 0.1} = T^a_{0.9, 0.9} = 0.9$ and $T^a_{0.1, 0.9} = T^a_{0.9, 0.1} = 0.1$, $\forall a \in \{1,2 \}$.
    That is, for each arm, state $1$ is highly rewarding whereas state $2$ is less rewarding, and the state of each arm stays with a probability of $0.9$ and switches with a probability of $0.1$.
    \item Asymmetric reward: two arms share the same transition dynamics, but the reward probability states of the two arms are different. 
    Specifically, the reward probability states for arm $a_1$ are $s^1_t \in \{0.1, 0.9\}$, and for arm $a_2$ are $s^2_t \in \{0.4, 0.6\}$, respectively. 
    Therefore the two states of arm $a_1$ are more differentiating than those of arm $a_2$.
    The transition dynamics is governed by $T^1_{0.1, 0.1} = T^1_{0.9, 0.9} = T^2_{0.4,0.4} = T^2_{0.6,0.6} = 0.9$, and $T^1_{0.1, 0.9} = T^1_{0.9, 0.1} = T^1_{0.4,0.6} = T^2_{0.6,0.4} = 0.1$.
    In other words, both arms follow the same dynamics that stay with a probability of $0.9$ and switch with a probability of $0.1$.
    \item Asymmetric transition: two arms share the same reward probability states $s^a_t \in \{0.1, 0.9\}, \forall a \in \{1,2\}$, but the transition dynamics are different.
    For arm $a_1$: $T^1_{0.1, 0.1} = T^1_{0.9, 0.9} = 0.9$ and $T^1_{0.1, 0.9} = T^1_{0.9, 0.1} = 0.1$; for arm $a_2$, $T^2_{0.1, 0.1} = T^2_{0.9, 0.9} = T^2_{0.1, 0.9} = T^2_{0.9, 0.1} = 0.5$.
    That is, whereas the transition dynamics for arm $a_1$ is the same as previous scenarios, the transition dynamics for arm $a_2$ is random with equal stay and switch probabilities of $0.5$.
\end{itemize}
With this task design, the belief state updates in the dynamic two-armed bandit POMDP tasks are analytically tractable by computing the posterior probability of each arm being in state $1$ conditioned on the history: $b_t = (p(s^1_t=\theta^1_1|h_t), p(s^2_t=\theta^2_1|h_t))$.
This belief update is tractable using standard Bayesian inference.
In turn, the Bayes-optimal policy can be derived using the value iteration algorithm \cite{Sutton1998} by discretizing the 2-dimensional belief state space.
The discounted cumulative return is computed with a discount factor $\gamma=0.95$.

\paragraph{Stationary Tiger}
To exemplify sequential decision-making under uncertainty, we consider the classic POMDP Tiger task \cite{kaelbling1998planning}.
In the tiger environment, an agent chooses between two doors—one hiding a tiger (penalty=$-100$) and the other hiding a treasure (reward=$10$). 
The agent may additionally choose to pay a small penalty=$-1$ for the ``listen'' action to acquire noisy observations about the tiger's location, 
The reliability of the noisy observation is controlled by the parameter of observation accuracy. 
To succeed, the agent must maintain a belief about the tiger's location to decide which door to open. 
This simple yet powerful paradigm tests an agent's ability to balance information gathering with reward-seeking, making it an ideal benchmark for evaluating RL algorithms in POMDP.
We consider two difficulty levels by varying the observation accuracy for the ``listen'' action.
Specifically, we consider a simpler task variant where the observation accuracy is $0.8$ and a harder task variant where the observation accuracy is $0.7$.
For the Tiger tasks, the Bayes-optimal agent tracks the belief of the tiger's location, for instance, by computing the posterior probability of the tiger being on the left given the history, $b_t = p(\text{tiger at the left}|h_t)$, using standard Bayesian inference.
The Bayes-optimal policy can be derived using value iteration \cite{Sutton1998} by discretizing the 1-dimensional belief state space. 
The discounted cumulative return is computed with a discount factor $\gamma=0.95$.
Typically, as the observation accuracy decreases, the Bayes-optimal solution will require listening for more times as observations are noisier before the belief states crosses the decision threshold for the optimal agent to decide on which door to open.

\paragraph{Dynamic Tiger}
We extend the classic Tiger task to a dynamic version by allowing the tiger's location to change over time, following Markov transition dynamics. 
We denote the tiger location $s$ to be in one of the $2$ states $s \in \{L, R\}$, and the tiger location evolves according to an independent Markovian transition process independent of the action chosen: $p(s_{t+1} | s_{t}, a) = p(s_{t+1} | s_{t}) = T_{s_{t}, s_{t+1}} $.
Specifically, we choose $T_{L,L} = T_{R,R}=0.9$ and $T_{L,R} = T_{R,L}=0.1$.
In other words, at each time step, the tiger stays with a probability of $0.9$ and switches its location with a probability of $0.1$.
As the information reliability and relevance are further corrupted by the dynamic nature of the hidden state, the task becomes more challenging because the agent has to balance listening more to increase its confidence against making decisions earlier in case the information gathered so far becomes obsolete. 
This task design permits tractable belief updates by tracking the posterior probability of the tiger being on the left given the history, $b_t = p(\text{tiger at the left}|h_t)$, using standard Bayesian inference incorporating the Markovian transition matrix.
The Bayes-optimal policy can be derived using value iteration \cite{Sutton1998} by discretizing the 1-dimensional belief state space. 
The discounted cumulative return is computed with a discount factor $\gamma=0.95$.
Similarly to the stationary Tiger tasks, we consider two difficulty levels by varing the observation accuracy to be $0.8$ or $0.7$.

\paragraph{Oracle bandit task} 
The oracle bandit task is designed to exemplify an environment where a successful policy requires paying an immediate exploration cost and utilize the information acquired to improve long-term return. 
In an 11-arm bandit environment with an episode length of 6 time steps, one of the first ten arms $a_{1-10}$ is selected uniformly randomly as the target arm $a^*$, which will give a payout of 5 upon choosing. 
The other nine arms out of $a_{1-10}$ are non-target arms, each of which will give a payout of 1 upon choosing. 
The last arm, $a_{11}$, is the oracle arm whose payout informs the index of the target arm in the form of $1/10$ of the target arm index $a^*$, i.e. $r(a_{11}) = 0.1*a^*$ (e.g. a reward of 0.3 from $a_{11}$ indicates that $a_3$ is the target arm). 
The average reward from the oracle arm is $0.55$ which is smaller than the payout from the non-target arms.
As a result, choosing the oracle arm is less favorable in terms of the immediate reward but provides useful information if an agent knows how to utilize to improve long-term return.
The discounted cumulative return is computed with a discount factor $\gamma=0.95$. 
Note that this setting is similar to a task considered in \citet{wang2016learning} but differs in that in their setup the oracle information was provided using a structured one-hot encoding format, but in our formulation no other feedback than the reward itself is given, which makes learning a good representation of the history more challenging and critical.
With knowledge of the task structure, the Bayes-optimal belief state is to track and update the posterior probability of each of the first ten arms being the true target arm conditioned on the history, i.e. $b_t = (p(a^*=a_1|h_t), p(a^*=a_2|h_t), ..., p(a^*=a_{10}|h_t))$.
The Bayes-optimal policy is to pull the oracle arm at the beginning and continue pulling the target arm as informed by the reward information from the oracle arm until the end of the episode. 

\paragraph{Latent goal cart}
Deriving Bayes-optimal solutions in continuous POMDPs are usually intractable, hindering rigorous representational equivalence analysis.
To enable evaluation in continuous observation and action spaces, we design an exemplar continuous control task which still permits tractable Bayes-optimal belief inference and policy.
In this task, an agent controls a continuous action, the velocity of a cart, to move along a 1-dimensional track to a hidden goal (\(+1\) or \(-1\)) which needs to be inferred from the continuous observation (current position) and reward (negative noisified distance from the hidden goal, with the noise following a Gaussian distribution) it receives.
We consider an episode length of 30 time steps.
This task design allows for tractable belief updates by tracking the posterior probability of the hidden goal being at \( +1 \) given the history, \( b_t = p(\text{goal at } +1 | h_t) \), using standard Bayesian inference incorporating the knowledge that the noise in reward follows a Gaussian distribution.
The Bayes-optimal policy can be derived using value iteration \cite{Sutton1998} by discretizing the 1-dimensional belief state space. 
The discounted cumulative return is computed with a discount factor $\gamma=0.95$.
This task is motivated by the Half-Cheetah-Dir task in MuJoCo, and can be seen as a simplified version to allow for tractable belief updates and value iteration to derive the Bayes-optimal solution.

\subsection{Agent details}
\label{agent_details}

\paragraph{RL\textsuperscript{2}}
The implementation of the baseline black-box memory-based meta-RL models, RL\textsuperscript{2}, follows an actor-critic architecture as in previous literature \cite{duan2016rl, wang2016learning}. 
Here we use RNNs that function as a memory module, consisting of $256$ hidden units and hyperbolic tangent activation functions. 
As shown in Fig. \ref{fig:intro}B, the input includes the current observation $o_t$, the previous action in one-hot format $a_{t-1}$, and the associated reward in scalar format $r_t$.
The recurrent layer is followed by a fully connected bottleneck layer designed to be the counterpart to the latent belief layer $b_t$ in Fig.\ref{fig:intro}C.
After the bottleneck layer is a hidden layer of $32$ units and a readout layer that generates a vector of logits for each action $a_t$ for the actor (or a vector that defines the mean and the standard deviation of a continuous Gaussian policy if solving a continuous task such as the Latent goal cart) and a scalar value baseline $V_t$ for the critic. 
Actions are then sampled from the softmax distribution defined by the action logits. 
The network is trained end-to-end to maximize the discounted cumulative reward with the Advantage Actor Critic algorithm \cite{mnih2016asynchronous} (Parts of the implementation code are based on \cite{pytorchrl}, under MIT license). 
The gradient of the objective function is given by:
\begin{equation} 
\label{formula:a2c_loss}
    \begin{split}
        \nabla \mathcal{L}_{A2C} 
        &= \nabla \mathcal{L}_\pi + \nabla \mathcal{L}_V + \nabla \mathcal{L}_{entropy} \\
        &= \frac{\partial \log \pi(a_t | \tau_{:t}; \psi)}{\partial \psi} \delta_t(\tau_{:t}; \psi_V) + \beta_V \delta_t(\tau_{:t}; \psi_V) \frac{\partial V}{\partial \psi_V} + \beta_e \frac{\partial H(\pi(a_t | \tau_{:t}; \psi))}{\partial \psi} \\
    \end{split}
\end{equation}
where
\begin{equation} 
\label{formula:advantage}
    \begin{split}
        \delta_t(\tau_{:t}; \psi_V) &= R_t - V(\tau_{:t}; \psi_V) \\
        R_t &= \sum_{i=0}^{k-1} [ \gamma^i r_{t+i} + \gamma^k V(\tau_{:t+k}; \psi_V) ]
    \end{split}
\end{equation}
defines the $n$-step temporal difference error advantage function $\delta_t$, the discounted $n$-step bootstrapped return $R_t$ with discount factor $\gamma$, $k$ the number of remaining time steps in the current episode, and $V$ the value function parametrized by $\psi_V$.
The neural network policy is denoted as $\pi$ and parametrized by $\psi$, and $H_\pi$ is the entropy of the policy.
Finally, $\beta_V$ and $\beta_e$ are hyperparemeters for controlling the relative weighting of value estimation loss and entropy regularization. 
The choice of hyperparameters for each task is summarized in Table. \ref{tab:model_hyperparameters}.
The neural network parameters are trained via backpropagation through time using the Adam Optimizer with a learning rate of $5\text{e-}5$.

\paragraph{Meta-RL with predictive modules} 
The self-supervised predictive modules are formulated as a VAE. 
For the encoder $q_\phi$ we use an RNN with $256$ hidden units and hyperbolic tangent activation functions. 
The output of the encoder RNN, i.e. the bottleneck layer, is treated as estimating the mean and the variance of the latents $m_t$, as standard in VAE. 
Therefore, the latent dimension is half of the bottleneck layer size.
The decoders $R_\theta$ and $T_\theta$ are multi-layered perceptrons (MLPs) with one hidden layer of $32$ units and ReLU activation functions. 
As shown in Fig. \ref{fig:intro}C, the encoder-decoder framework takes as input the current observation $o_t$, the previous action in one-hot format $a_{t-1}$, and the associated scalar reward $r_t$, and is set up to make prediction of the upcoming observations $o_{t+1}$ and rewards $r_{t+1}$, conditioned on the trajectory as incurred by the policy network $\pi_\psi$ described below. 
The entire VAE is trained to maximize the ELBO (Eq. \ref{formula:elbo}) as derived in \ref{elbo_derivation}. 
A coefficient of $0.01$ is used for the relative contribution of the KL-term for training the VAE.
We use the Adam Optimizer with a learning rate of $7\text{e-}5$ to train the VAE using backpropagation through time.

The policy network $\pi_\psi$ is parametrized as an MLP with one hidden layer of $32$ units and hyperbolic tangent activation functions.
Similar to the previous paragraph on RL\textsuperscript{2}, the policy network is trained with the Advantage Actor Critic algorithm to optimize the same loss function as described in Eq. \ref{formula:a2c_loss}. 
The choice of hyperparameters for each task is summarized in Table. \ref{tab:model_hyperparameters}.
An Adam Optimizer with a learning rate of $5\text{e-}5$ is used to optimize the policy network. 
Note although the policy loss depends on the parameters of the encoder $q_\phi$, we do not backpropagate the policy loss gradient through the encoder as the goal of the encoder is to learn a belief $b_t$ over the latent states predictive of the future such that the belief alone should be a sufficient representation for policy learning (similar to \citet{zintgraf2021varibad}). 
Parts of the implementation code are based on \cite{zintgraf2021varibad} (under MIT license).

\paragraph{Model training} The above meta-RL models (RL\textsuperscript{2} and ours, meta-RL with predictive modules) are trained on internal GPU clusters (NVIDIA GeForce RTX 4090), which takes less than 1G GPU memory and between 10-48 hours for training per model, depending on the task.
Compared to RL\textsuperscript{2}, training time of meta-RL with predictive modules increases by \(\sim\) 30-40\% with the overhead coming from two additional decoders and VAE training, whereas inference time is comparable.

\paragraph{Ablation study} 
To pinpoint which algorithmic design choices drive enhanced representation learning in our proposed meta-RL with predictive modules, we performed two targeted ablations: 
\begin{itemize}
    \item (i) no KL: where we remove the VAE’s KL regularization to test whether latent regularization is necessary to enforce compact representations.
    \item (ii) joint RL: where we allow the policy gradients from RL loss to jointly train the RNN encoder alongside the predictive loss---rather than training the encoder \emph{solely} with the predictive loss---to assess whether policy gradients provide additional representational benefits.
\end{itemize}

\begin{table}
  \caption{\textbf{Hyperparameters} 
  }
  \label{tab:model_hyperparameters}
  \footnotesize
  \setlength{\tabcolsep}{2.5pt}
  \centering
  \begin{tabular}{lccccc}
    \toprule
     & episode length & $\beta_e$ & $\beta_{V}$ & n\_updates & bottleneck size \\
    Two-armed bandit & $40$ & $0.01-0.05$ & $0.01-0.05$ & $3e5$ & $8$ \\
    Dynamic bandit & $300$ & $0.01$ & $0.05$ & $1e5$ & $8$ \\
    Stationary Tiger & $20-30$ & $0.3$ & $0.1$ & $3e5$ & $4$ \\
    Dynamic Tiger & $30-40$ & $0.3$ & $0.1$ & $3e5$ & $4$ \\
    Oracle bandit & $6$ & $0.3-0.5$ & $0.01-0.05$ & $3e6$ & $16$ \\
    Latent goal cart & $30$ & $0.005-0.01$ & $0.01$ & $3e5$ & $8$ \\
    \bottomrule
    
  \end{tabular}
\end{table}

\subsection{State machine simulation}
\label{sma}

Following the procedure of state machine simulation introduced in \citet{mikulik2020meta}, we consider whether a meta-RL system (RL\textsuperscript{2} or our proposed approach, meta-RL with predictive modules) can both \emph{simulate} and \emph{be simulated by}, a Bayes-optimal agent for a given POMDP task. 

To evaluate how well a state machine $M$ \emph{simulates} another machine $N$, a function $\phi$ is first learned to map the states $S_N$ in $N$ into the state space $S_M$ of $M$. 
As enumeration over all possible trajectories are not practical if not possible, quality of a simulation is measured along trajectories sampled from some reference distribution. 
Given trajectories from a reference distribution, quality of the simulation is then measured by (i) the \emph{state-transition dissimilarity} $D_s$, measured as the mean-squared error (MSE) between the embedded states $\phi(S_N)$ and the target states $S_M$, and (ii) the \emph{output dissimilarity} $D_o$, measured as the difference in the expected return generated from the states $S_N$ using the machine $N$ and those generated from the states $\phi(S_N)$ using the machine $M$. If both the state-transition and output dissimilarities $D_s$ and $D_o$ are low/ negligible, then we establish that $M$ simulates $N$. If both $M$ simulates $N$ and $N$ simulates $M$, then we can say $M$ and $N$ are computationally equivalent, and their states $S_N$ and $S_M$ are equivalent.

In practice, the mapping function $\phi$ is implemented as an MLP with ReLU activations and three hidden layers of 64, 128, and 64 units, respectively. 
The MLP is trained with the Adam Optimizer with learning rate $0.001$ and batch size $64$. The training set is consisted of $300-1000$ trajectories depending on the variability of each task distribution, and the results reported are from another test set of $300-1000$ trajectories. 
Compared with \citet{mikulik2020meta}, where the reference distribution is generated by the meta-RL agent, here we modify the procedure by generating the reference distribution using the Bayes-optimal agent, as this provides an even more stringent condition where the simulation is evaluated in the regime of Bayes-optimal solutions.

When evaluating how well a Bayes-optimal agent \emph{simulates} a meta-RL system, we first train an MLP mapping meta-RL states into Bayes-optimal states by minimizing the MSE between the mapped states and the target Bayes-optimal states. 
After training, quality of the simulation is measured on a test set by evaluating the state-transition dissimilarity ($D_s$: metaRL$\rightarrow$Bayes) and the output dissimilarity ($D_o$: metaRL$\rightarrow$Bayes). If both $D_s$: metaRL$\rightarrow$Bayes and $D_o$: metaRL$\rightarrow$Bayes are low after training, then the Bayes-optimal agent \emph{simulates} the meta-RL agent.

On the other hand, to evaluate how well a Bayes-optimal agent \emph{is simulated by} a meta-RL one, an MLP is trained to map Bayes-optimal states into meta-RL states, and the state-transition dissimilarity is denoted $D_s$: Bayes$\rightarrow$metaRL and the output dissimilarity denoted $D_o$: Bayes$\rightarrow$metaRL. 
If both $D_s$: Bayes$\rightarrow$metaRL and $D_o$: Bayes$\rightarrow$metaRL are low after training, then the Bayes-optimal agent \emph{is simulated by} the meta-RL agent.

If all the above four dissimilarity measures are low/ negligible, then we can say that the meta-RL agent and the Bayes-optimal agent are computationally equivalent and their representations are equivalent. 
Our results in this paper demonstrate that after training the proposed meta-RL with predictive coding modules can attain much lower dissimilarities than conventional RL\textsuperscript{2}, indicating that meta-RL with predictive modules can more closely approximate the Bayes-optimal belief states. 

Note that before training, the state dissimilarity $D_s$ can be low for the recurrent layers in all models, similar to what \citet{mikulik2020meta} reported and discussed---untrained RNNs may maintain a verbose representation of the history by embedding each trajectory into a unique hidden state, which can be subsequently mapped to minimally sufficient Bayes-optimal statistic with low error using expressive enough functions, like the MLPs used in the above state machine simulation procedure. 
This observation also highlights that using decoding alone may not provide a thorough assessment of representation equivalence, and we need to consider their structural and computational relevance when comparing representations.

\newpage
\subsection{Additional results}

\subsubsection{State space visualization}
\label{append_state_space_visualization}
\begin{figure}[h!]
  \centering
  \includegraphics[width=1.0\textwidth]{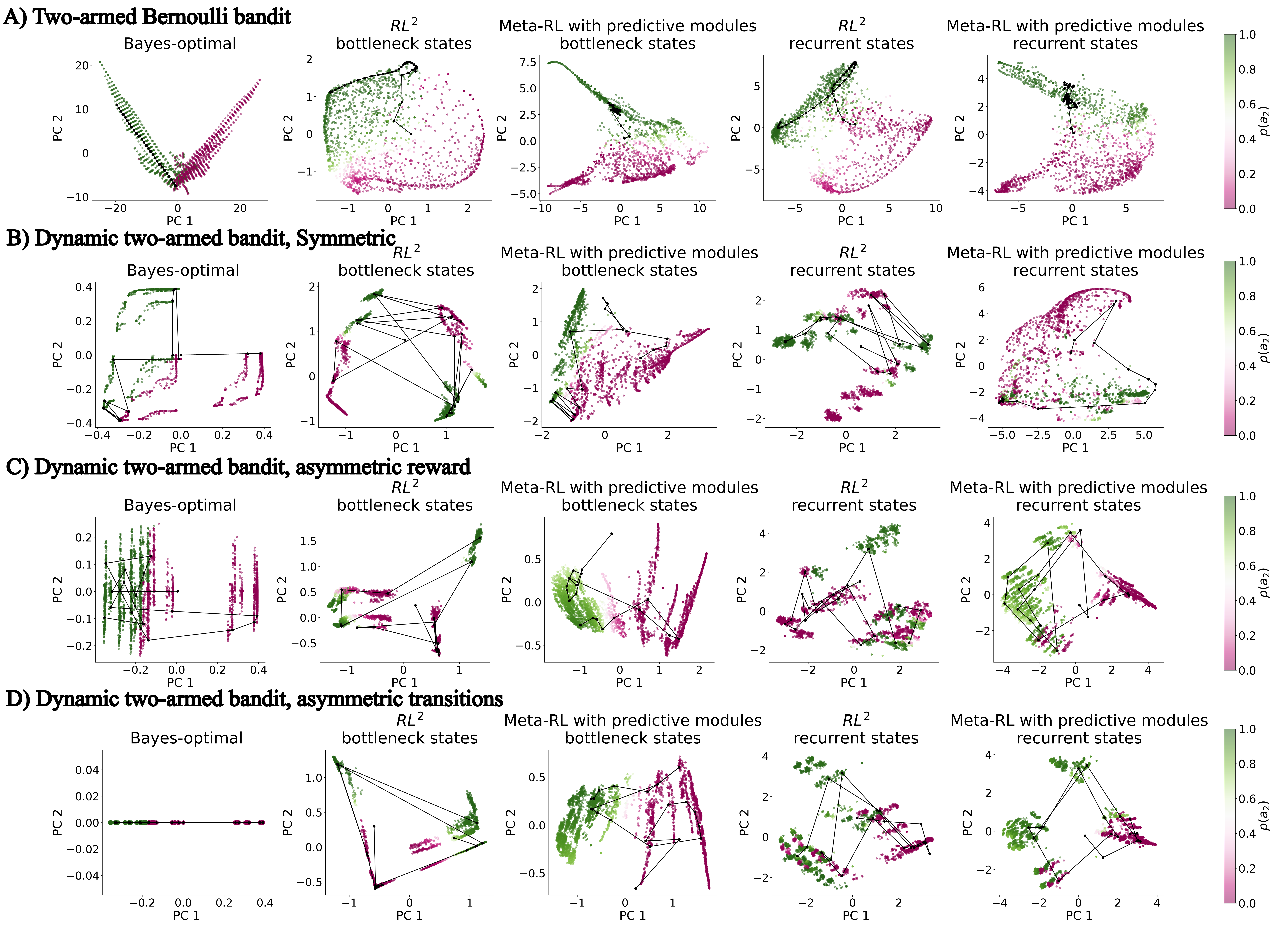}
  \caption{\textbf{Visualization of state space in the stationary and dynamic two-armed bandit tasks}.
  Example state space for A) Two-armed Bernoulli bandit task, B) Dynamic two-armed bandit task with symmetric reward and transition, C) Dynamic two-armed bandit task with asymmetric reward, and D) Dynamic two-armed bandit task with asymmetric transition.
  For each panel, the first two principal components are plotted.
  States are colored by the corresponding policy, i.e. probability of choosing $a_2$.
  Black curves show one example trajectory.
  For each task, from the left to the right are the state space of the Bayes-optimal agent (i.e. the belief states), the bottleneck layer in the RL\textsuperscript{2} model, the bottleneck layer in the meta-RL with predictive modules, the recurrent layer in the RL\textsuperscript{2} model, and the recurrent layer in the meta-RL with predictive modules, respectively.
  }
  \label{fig:append_state_space_bandits}
\end{figure}

\begin{figure}[h!]
  \centering
  \includegraphics[width=1.0\textwidth]{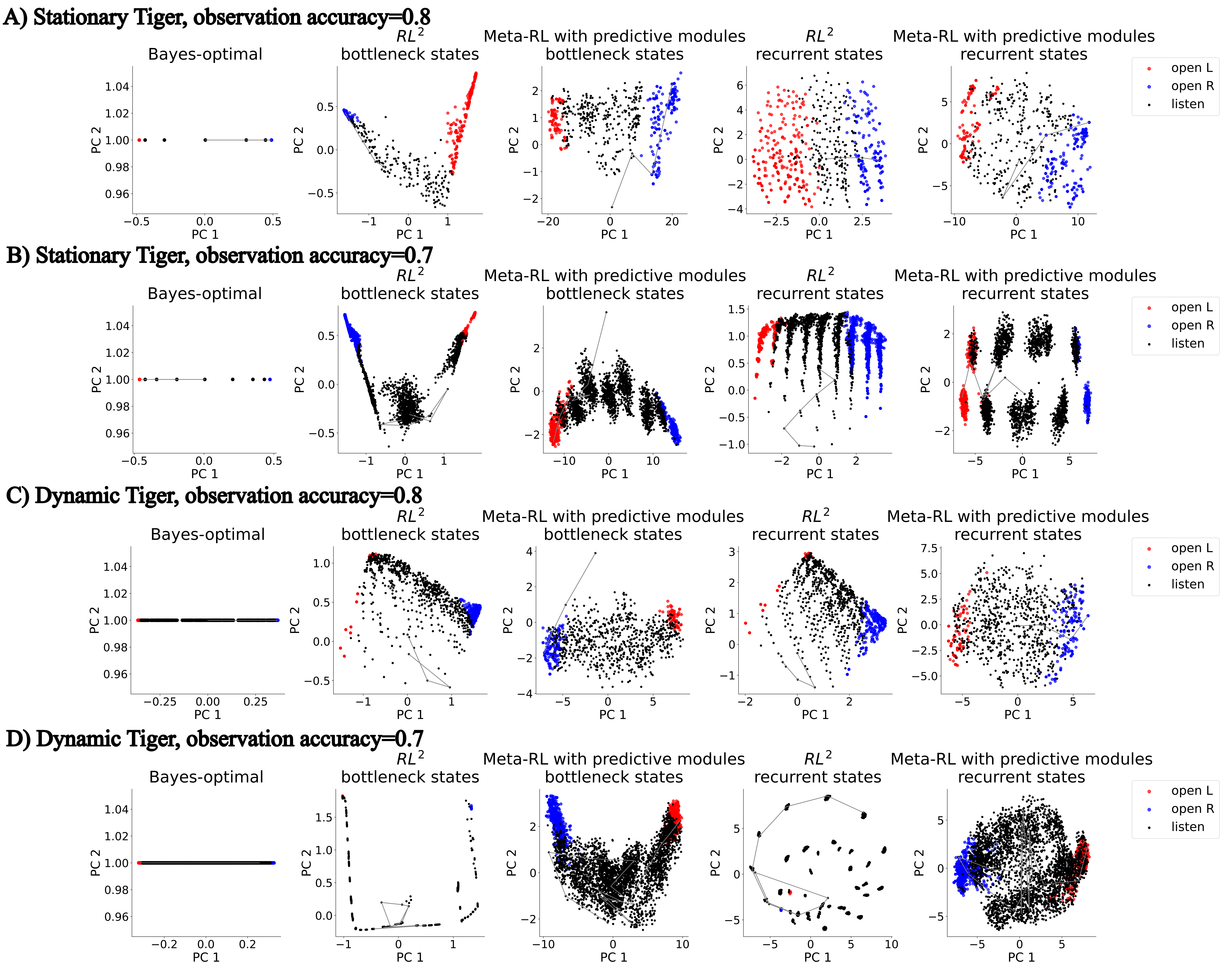}
  \caption{\textbf{Visualization of state space in the stationary and dynamic Tiger tasks}.
  Example state space for A) Stationary Tiger task with observation accuracy of $0.8$, B) Stationary Tiger task with observation accuracy of $0.7$, C) Dynamic Tiger task with observation accuracy of $0.8$, and D) Dynamic Tiger task with observation accuracy of $0.7$.
  For each panel, the first two principal components are plotted.
  States are colored by the corresponding policy, with black denoting choosing listen, red choosing to open the left door, and blue choosing to open the right door.
  Black curves show one example trajectory.
  For each task, from the left to the right are the state space of the Bayes-optimal agent (i.e. the belief states), the bottleneck layer in the RL\textsuperscript{2} model, the bottleneck layer in the meta-RL with predictive modules, the recurrent layer in the RL\textsuperscript{2} model, and the recurrent layer in the meta-RL with predictive modules, respectively.
  }
  \label{fig:append_state_space_tigers}
\end{figure}

\begin{figure}[h!]
  \centering
  \includegraphics[width=1.0\textwidth]{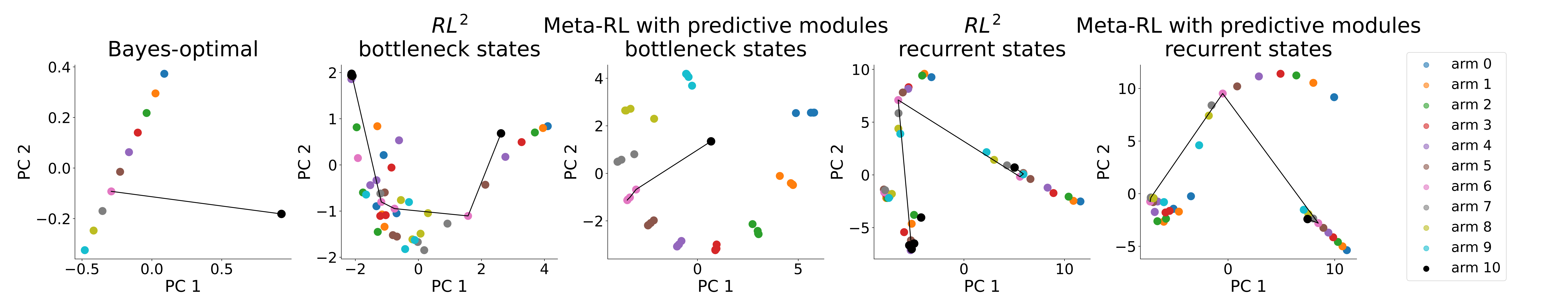}
  \caption{\textbf{Visualization of state space in the oracle bandit task}.
  Example state space for the oracle bandit task.
  For each panel, the first two principal components are plotted.
  States are colored by the corresponding policy, with black denoting choosing the oracle arm $a_{11}$ and other colors denoting choosing one of the first ten arms $a_{1-10}$.
  Black curves show one example trajectory.
  From the left to the right are the state space of the Bayes-optimal agent (i.e. the belief states), the bottleneck layer in the RL\textsuperscript{2} model, the bottleneck layer in the meta-RL with predictive modules, the recurrent layer in the RL\textsuperscript{2} model, and the recurrent layer in the meta-RL with predictive modules, respectively.
  }
  \label{fig:append_state_space_oracle}
\end{figure}

\begin{figure}[t!]
  \centering
  \includegraphics[width=1.0\textwidth]{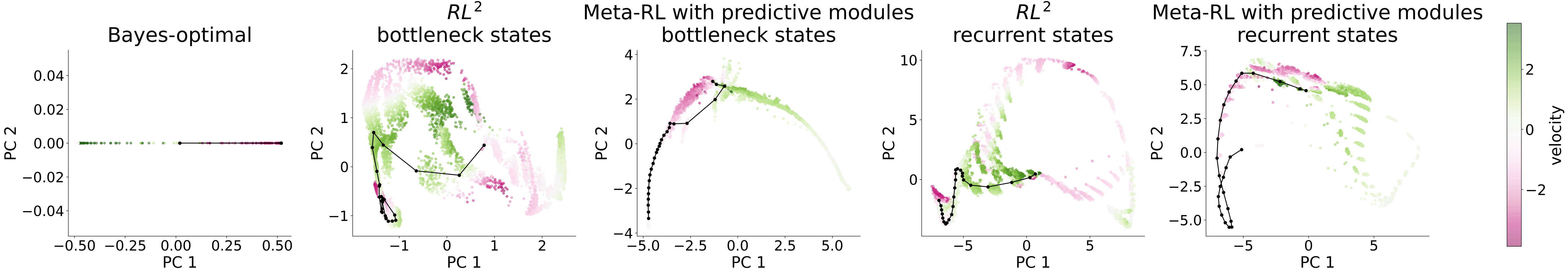}
  \caption{\textbf{Visualization of state space in the latent goal cart task}.
  Example state space for the oracle bandit task.
  For each panel, the first two principal components are plotted.
  States are colored by the corresponding policy, i.e. the velocity of the cart.
  Black curves show one example trajectory.
  From the left to the right are the state space of the Bayes-optimal agent (i.e. the belief states), the bottleneck layer in the RL\textsuperscript{2} model, the bottleneck layer in the meta-RL with predictive modules, the recurrent layer in the RL\textsuperscript{2} model, and the recurrent layer in the meta-RL with predictive modules, respectively.
  }
  \label{fig:append_state_space_latent_goal_cart}
\end{figure}

To understand the structure of the learned representations in meta-RL agents and qualitatively compare those to the Bayes-optimal belief states, visualization of example state spaces for the stationary and dynamic bandit tasks, the stationary and dynamic Tiger tasks, the oracle bandit tasks, and the latent goal cart tasks are presented in Figures. \ref{fig:append_state_space_bandits}, \ref{fig:append_state_space_tigers}, \ref{fig:append_state_space_oracle}, and \ref{fig:append_state_space_latent_goal_cart} respectively.

\clearpage
\subsubsection{Full state machine simulation results}
\label{append_sms_full_results}

\begin{figure}[ht!]
  \centering
  \includegraphics[width=1.0\textwidth]{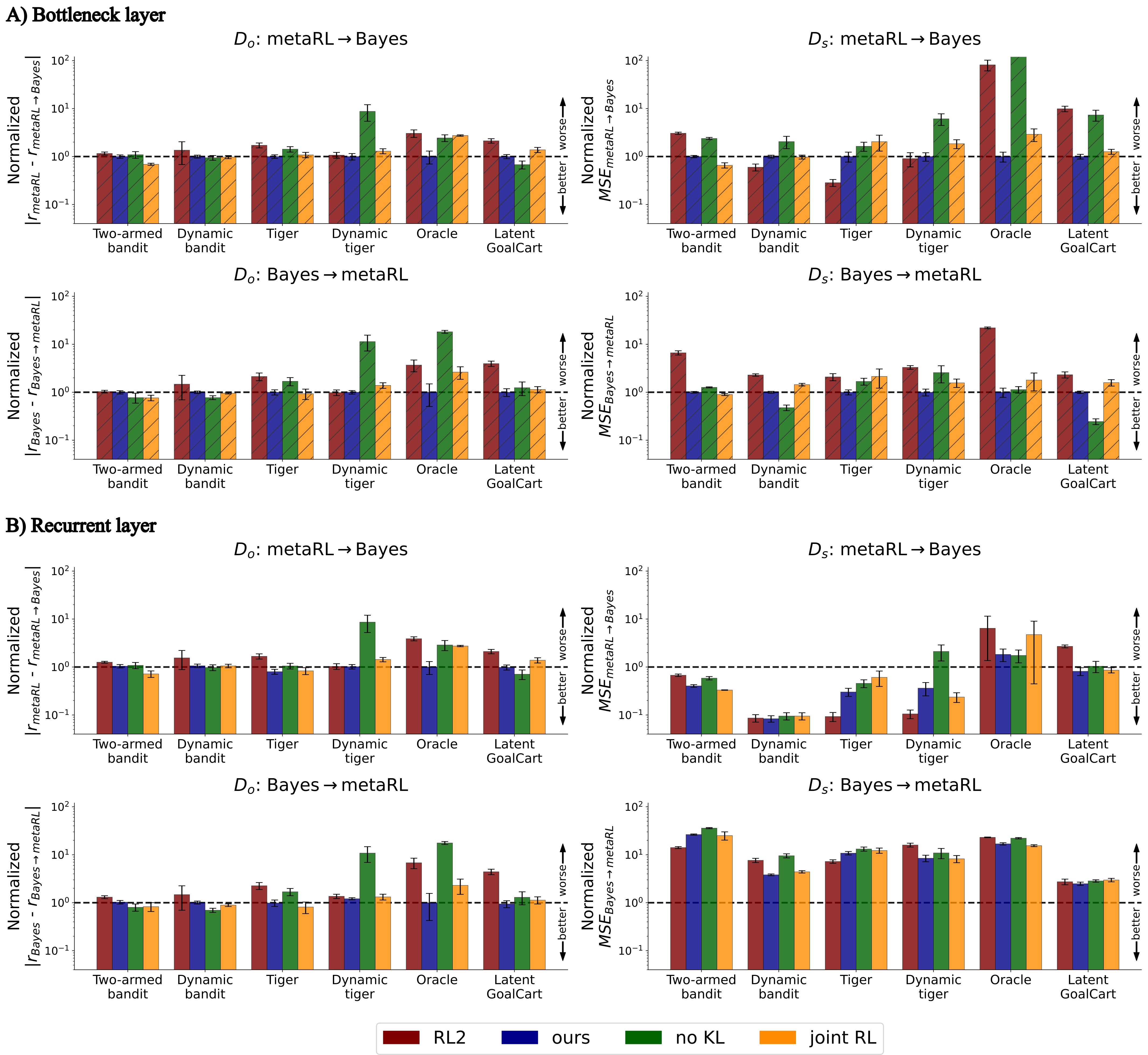}
  \caption{\textbf{Results of state machine simulation}.
  State and output dissimilarities ($D_o$ and $D_s$) for both mapping directions (meta-RL$\rightarrow$Bayes and Bayes$\rightarrow$meta-RL) across all tasks considered, using A) the bottleneck layer and B) the recurrent layer of the fully-trained models of RL\textsuperscript{2} (red), meta-RL with predictive modules (ours, blue), no KL (where KL regularization is ablated for VAE training, green), and joint RL (where the policy gradient of RL loss is used to jointly train the RNN encoder with the predictive loss, orange).
  To pool across different variants of the same task type, all dissimilarity measures are normalized by the mean of the corresponding dissimilarities of the bottleneck layer in the meta-RL with predictive modules models (denoted as \emph{ours} in the legend.)
  In other words, when normalized with the mean bottleneck dissimilarity of our model, values significantly greater than $1$ indicate that the representation is worse than the bottleneck layer in our model (i.e. deviating more from the Bayes-optimal belief states). 
  In contrast, values significantly smaller than $1$ indicate the representation is better than the bottleneck layer in our model (i.e. more closely approximating the Bayes-optimal belief states).
  Error bars denote s.e.m. across at least five random seeds per model type.
  }
  \label{fig:append_sms_full}
\end{figure}

Full results of state machine simulation analysis on the trained models of RL\textsuperscript{2} and the proposed meta-RL with predictive modules (ours), together with models considered in the ablation studies---no KL and joint RL, are summarized in Fig. \ref{fig:append_sms_full} and Tables \ref{tab:full_state_machine_bandits}, \ref{tab:full_state_machine_tigers}, \ref{tab:full_state_machine_oracle}, and
\ref{tab:full_state_machine_latent_goal_cart}.

\begin{table}[t]
  \caption{\textbf{State machine simulation results for the stationary and dynamic bandit tasks}: 
  State and output dissimilarities of the fully-trained models for both mapping directions. 
  $M$ for meta-RL, $B$ for Bayes-optimal solutions.
  Values indicate mean±s.t.d.
  Bold indicates the best dissimilarities among the trained models and better than untrained models ($p<0.05$).
  }
  \label{tab:full_state_machine_bandits}
  \footnotesize
  \setlength{\tabcolsep}{2.5pt}
  \centering
  \begin{tabular}{lcccccccc}
    
    \toprule
    \multicolumn{9}{c}{\textbf{Two-armed Bernoulli bandit} (unit: 1e-2)} \\
     & RL\textsuperscript{2} & ours & no KL & joint RL & RL\textsuperscript{2} & ours & no KL & joint RL \\
    \cmidrule(l){2-5} \cmidrule(l){6-9}
     & \multicolumn{4}{c}{Bottleneck} & \multicolumn{4}{c}{Recurrent} \\
    $D_o^{M \rightarrow B}$ &
    1.52±0.36 & 1.31±0.37 & 1.44±0.4 & \textbf{0.91±0.05} & 1.65±0.24 & 1.37±0.38 & 1.42±0.36 & \textbf{0.94±0.15}  \\
    $D_s^{M \rightarrow B}$ &
    5.41±0.96 & 1.77±0.33 & 4.21±0.38 & 1.16±0.15 & 1.19±0.23 & 0.71±0.16 & 1.04±0.14 & 0.58±0.01  \\
    $D_o^{B \rightarrow M}$ &
    \textbf{1.19±0.25} & \textbf{1.15±0.31} & \textbf{0.89±0.36} & \textbf{0.88±0.12} & \textbf{1.51±0.35} & \textbf{1.19±0.34} & \textbf{0.93±0.27} & \textbf{0.95±0.2}  \\
    $D_s^{B \rightarrow M}$ & 
    5.38±1.81 & \textbf{0.81±0.11} & 1.02±0.04 & \textbf{0.73±0.04} & 11.43±1.83 & 21.37±2.12 & 29.05±1.43 & 20.33±3.96  \\
    \midrule

    \multicolumn{9}{c}{\textbf{Dynamic two-armed bandit: symmetric reward and transition} (unit: 1e-2)} \\
     & RL\textsuperscript{2} & ours & no KL & joint RL & RL\textsuperscript{2} & ours & no KL & joint RL \\
    \cmidrule(l){2-5} \cmidrule(l){6-9}
     & \multicolumn{4}{c}{Bottleneck} & \multicolumn{4}{c}{Recurrent} \\
    $D_o^{M \rightarrow B}$ &
    \textbf{2.09±3.95} & \textbf{1.17±0.18} & \textbf{0.86±0.48} & \textbf{1.06±0.1} & \textbf{2.18±3.92} & \textbf{1.25±0.21} & \textbf{0.86±0.43} & \textbf{1.08±0.12} \\
    $D_s^{M \rightarrow B}$ &
    0.19±0.02 & 0.21±0.05 & 0.49±0.36 & 0.22±0.1 & 0.03±0.01 & 0.03±0.01 & 0.03±0.02 & 0.04±0.01 \\
    $D_o^{B \rightarrow M}$ &
    \textbf{2.06±3.81} & \textbf{0.98±0.36} & \textbf{0.79±0.33} & \textbf{0.89±0.19} & \textbf{2.18±3.75} & \textbf{1.05±0.44} & \textbf{0.72±0.3} & \textbf{0.82±0.3} \\
    $D_s^{B \rightarrow M}$ & 
    2.38±0.6 & 1.09±0.22 & \textbf{0.45±0.13} & 1.43±0.23 & 10.26±2.01 & 4.86±0.44 & 10.25±2.37 & 6.11±0.69 \\
    \midrule

    \multicolumn{9}{c}{\textbf{Dynamic two-armed bandit: asymmetric reward} (unit: 1e-2)} \\
     & RL\textsuperscript{2} & ours & no KL & joint RL & RL\textsuperscript{2} & ours & no KL & joint RL \\
    \cmidrule(l){2-5} \cmidrule(l){6-9}
     & \multicolumn{4}{c}{Bottleneck} & \multicolumn{4}{c}{Recurrent} \\
    $D_o^{M \rightarrow B}$ &
    \textbf{0.8±0.28} & \textbf{0.86±0.27} & \textbf{0.76±0.19} & \textbf{0.82±0.24} & \textbf{1.01±0.14} & \textbf{1.22±0.34} & \textbf{0.59±0.13} & \textbf{1.04±0.34} \\
    $D_s^{M \rightarrow B}$ &
    0.34±0.06 & 0.93±0.11 & 0.4±0.17 & 1.04±0.16 & 0.04±0.01 & 0.05±0.0 & 0.03±0.01 & 0.04±0.01 \\
    $D_o^{B \rightarrow M}$ &
    \textbf{1.15±0.27} & \textbf{1.39±0.03} & \textbf{0.99±0.37} & \textbf{1.37±0.16} & \textbf{1.03±0.31} & \textbf{1.37±0.03} & \textbf{0.97±0.34} & \textbf{1.28±0.35} \\
    $D_s^{B \rightarrow M}$ & 
    3.49±0.51 & \textbf{1.45±0.1} & \textbf{0.99±0.44} & 1.98±0.33 & 10.47±3.6 & 4.91±0.75 & 18.2±5.4 & 5.56±0.6 \\
    \midrule

    \multicolumn{9}{c}{\textbf{Dynamic two-armed bandit: asymmetric transition} (unit: 1e-2)} \\
     & RL\textsuperscript{2} & ours & no KL & joint RL & RL\textsuperscript{2} & ours & no KL & joint RL \\
    \cmidrule(l){2-5} \cmidrule(l){6-9}
     & \multicolumn{4}{c}{Bottleneck} & \multicolumn{4}{c}{Recurrent} \\
    $D_o^{M \rightarrow B}$ &
    \textbf{0.85±0.23} & \textbf{0.93±0.33} & \textbf{1.11±0.47} & \textbf{0.98±0.39} & \textbf{1.22±0.33} & \textbf{0.78±0.29} & \textbf{1.39±0.56} & \textbf{0.95±0.4} \\
    $D_s^{M \rightarrow B}$ &
    0.04±0.02 & 0.33±0.11 & 1.1±1.06 & 0.24±0.09 & 0.01±0.0 & 0.01±0.01 & 0.03±0.01 & 0.02±0.01 \\
    $D_o^{B \rightarrow M}$ &
    \textbf{1.52±0.1} & \textbf{1.7±0.17} & \textbf{1.36±0.46} & \textbf{1.67±0.12} & \textbf{1.21±0.39} & \textbf{1.64±0.19} & \textbf{1.12±0.5} & \textbf{1.55±0.16} \\
    $D_s^{B \rightarrow M}$ & 
    5.14±0.75 & 2.15±0.45 & \textbf{0.7±0.39} & 3.56±1.08 & 9.0±1.06 & 7.41±0.91 & 14.35±5.85 & 8.16±1.69 \\
    \bottomrule

  \end{tabular}
\end{table}

\begin{table}
  \caption{\textbf{State machine simulation results for the stationary and dynamic Tiger tasks}.
  }
  \label{tab:full_state_machine_tigers}
  \footnotesize
  \setlength{\tabcolsep}{2.5pt}
  \centering
  \begin{tabular}{lcccccccc}
    
    \toprule
    
    \multicolumn{9}{c}{\textbf{Stationary Tiger: observation accuracy 0.8} (unit: 1e-2)} \\
     & RL\textsuperscript{2} & ours & no KL & joint RL & RL\textsuperscript{2} & ours & no KL & joint RL \\
    \cmidrule(l){2-5} \cmidrule(l){6-9}
     & \multicolumn{4}{c}{Bottleneck} & \multicolumn{4}{c}{Recurrent} \\
    $D_o^{M \rightarrow B}$ &
    \textbf{3.33±0.91} & \textbf{2.57±1.01} & \textbf{2.93±1.09} & \textbf{2.97±0.87} & \textbf{3.05±0.64} & \textbf{2.07±1.03} & \textbf{2.13±0.37} & \textbf{2.6±1.22} \\
    $D_s^{M \rightarrow B}$ &
    0.08±0.05 & 0.31±0.33 & 0.46±0.3 & 0.67±0.87 & 0.03±0.03 & 0.06±0.04 & 0.11±0.04 & 0.23±0.27 \\
    $D_o^{B \rightarrow M}$ &
    \textbf{3.03±1.54} & \textbf{2.26±0.88} & \textbf{2.98±1.7} & \textbf{1.88±1.24} & \textbf{4.15±1.72} & \textbf{1.85±1.22} & \textbf{2.98±1.7} & \textbf{1.34±0.92} \\
    $D_s^{B \rightarrow M}$ & 
    8.04±4.38 & \textbf{2.92±1.84} & 4.1±1.3 & 9.07±10.54 & 24.95±7.07 & 28.31±8.99 & 35.61±8.74 & 35.95±12.55 \\
    \midrule

    \multicolumn{9}{c}{\textbf{Stationary Tiger: observation accuracy 0.7} (unit: 1e-2)} \\
     & RL\textsuperscript{2} & ours & no KL & joint RL & RL\textsuperscript{2} & ours & no KL & joint RL \\
    \cmidrule(l){2-5} \cmidrule(l){6-9}
     & \multicolumn{4}{c}{Bottleneck} & \multicolumn{4}{c}{Recurrent} \\
    $D_o^{M \rightarrow B}$ &
    3.25±1.13 & \textbf{1.57±0.49} & 2.68±1.0 & \textbf{1.58±0.74} & 3.27±1.14 & \textbf{1.24±0.62} & 2.0±0.9 & \textbf{1.02±0.36} \\
    $D_s^{M \rightarrow B}$ &
    0.09±0.05 & 0.3±0.27 & 0.53±0.39 & 0.58±0.44 & 0.03±0.01 & 0.13±0.09 & 0.17±0.1 & 0.15±0.08 \\
    $D_o^{B \rightarrow M}$ &
    \textbf{3.06±1.54} & \textbf{1.09±0.73} & \textbf{2.25±1.34} & \textbf{1.13±0.85} & \textbf{2.86±1.67} & \textbf{1.33±0.76} & \textbf{2.25±1.1} & \textbf{1.13±0.85} \\
    $D_s^{B \rightarrow M}$ & 
    4.51±0.96 & \textbf{2.97±0.78} & 5.81±3.41 & \textbf{3.46±1.53} & 18.18±1.76 & 36.56±11.68 & 42.07±15.13 & 36.21±14.37 \\
    \midrule

    \multicolumn{9}{c}{\textbf{Dynamic Tiger: observation accuracy 0.8} (unit: 1e-2)} \\
     & RL\textsuperscript{2} & ours & no KL & joint RL & RL\textsuperscript{2} & ours & no KL & joint RL \\
    \cmidrule(l){2-5} \cmidrule(l){6-9}
     & \multicolumn{4}{c}{Bottleneck} & \multicolumn{4}{c}{Recurrent} \\
    $D_o^{M \rightarrow B}$ &
    \textbf{2.19±0.96} & \textbf{2.39±0.41} & \textbf{3.5±1.35} & \textbf{3.48±1.18} & \textbf{2.25±0.95} & \textbf{2.16±0.7} & \textbf{3.1±0.22} & \textbf{3.93±1.03} \\
    $D_s^{M \rightarrow B}$ &
    0.04±0.01 & 0.1±0.05 & 0.31±0.17 & 0.29±0.06 & 0.01±0.0 & 0.04±0.04 & 0.12±0.09 & 0.02±0.02 \\
    $D_o^{B \rightarrow M}$ &
    \textbf{2.01±0.93} & \textbf{1.99±0.45} & 3.48±0.6 & 3.28±1.44 & 3.28±0.59 & \textbf{2.34±0.42} & 3.21±0.76 & 3.28±1.26 \\
    $D_s^{B \rightarrow M}$ & 
    6.27±1.15 & \textbf{1.78±0.98} & \textbf{2.9±1.76} & \textbf{2.03±0.88} & 29.5±2.07 & 14.39±4.93 & 24.2±10.58 & 11.82±5.04 \\
    \midrule

    \multicolumn{9}{c}{\textbf{Dynamic Tiger: observation accuracy 0.7} (unit: 1e-2)} \\
     & RL\textsuperscript{2} & ours & no KL & joint RL & RL\textsuperscript{2} & ours & no KL & joint RL \\
    \cmidrule(l){2-5} \cmidrule(l){6-9}
     & \multicolumn{4}{c}{Bottleneck} & \multicolumn{4}{c}{Recurrent} \\
    $D_o^{M \rightarrow B}$ &
    \textbf{2.51±0.66} & \textbf{2.05±1.28} & 27.92±21.09 & \textbf{2.33±1.01} & \textbf{2.3±0.67} & \textbf{2.37±0.53} & 27.75±21.84 & \textbf{2.51±0.95} \\ 
    $D_s^{M \rightarrow B}$ &
    0.39±0.23 & 0.28±0.17 & 2.23±1.53 & 0.2±0.18 & 0.03±0.02 & 0.08±0.06 & 0.74±0.77 & 0.07±0.06 \\
    $D_o^{B \rightarrow M}$ &
    \textbf{1.73±0.32} & \textbf{1.8±0.47} & 32.11±22.34 & \textbf{2.05±0.57} & \textbf{1.95±0.31} & \textbf{2.27±0.17} & 30.65±20.94 & \textbf{1.8±0.65} \\
    $D_s^{B \rightarrow M}$ & 
    7.1±1.89 & \textbf{2.3±0.72} & 7.37±8.41 & 4.62±2.85 & 35.44±12.21 & 20.65±10.63 & 21.05±19.62 & 22.64±12.53 \\
    \bottomrule
    
  \end{tabular}
\end{table}

\begin{table}
  \caption{\textbf{State machine simulation results for the oracle bandit tasks}.
  }
  \label{tab:full_state_machine_oracle}
  \footnotesize
  \setlength{\tabcolsep}{2.5pt}
  \centering
  \begin{tabular}{lcccccccc}
    
    \toprule
    \multicolumn{9}{c}{\textbf{Oracle bandit} (unit: 1e-2)} \\
     & RL\textsuperscript{2} & ours & no KL & joint RL & RL\textsuperscript{2} & ours & no KL & joint RL \\
    \cmidrule(l){2-5} \cmidrule(l){6-9}
     & \multicolumn{4}{c}{Bottleneck} & \multicolumn{4}{c}{Recurrent} \\
    $D_o^{M \rightarrow B}$ &
    17.21±9.73 & \textbf{5.64±4.5} & 13.79±4.44 & 15.48±1.02 & 21.96±6.65 & \textbf{5.64±4.5} & 16.26±7.76 & 15.48±1.02 \\
    $D_s^{M \rightarrow B}$ &
    6.45±5.14 & \textbf{0.08±0.05} & 24.4±6.94 & 0.23±0.14 & 0.51±1.27 & 0.14±0.11 & 0.14±0.08 & 0.37±0.68 \\
    $D_o^{B \rightarrow M}$ &
    12.01±10.45 & \textbf{3.26±4.26} & 59.34±8.43 & 8.57±5.0 & 22.18±17.23 & \textbf{3.24±4.9} & 57.34±8.39 & 7.49±5.26 \\
    $D_s^{B \rightarrow M}$ & 
    51.29±6.73 & \textbf{2.33±1.34} & 2.64±0.82 & 4.18±3.38 & 53.72±2.63 & 39.33±5.68 & 51.5±3.42 & 36.02±3.26 \\
    \bottomrule
    
  \end{tabular}
\end{table}

\begin{table}[t!]
  \caption{\textbf{State machine simulation results for the latent goal cart tasks}.
  }
  \label{tab:full_state_machine_latent_goal_cart}
  \footnotesize
  \setlength{\tabcolsep}{2.5pt}
  \centering
  \begin{tabular}{lcccccccc}
    
    \toprule
    \multicolumn{9}{c}{\textbf{Latent goal cart} (unit: 1e-2)} \\
     & RL\textsuperscript{2} & ours & no KL & joint RL & RL\textsuperscript{2} & ours & no KL & joint RL \\
    \cmidrule(l){2-5} \cmidrule(l){6-9}
     & \multicolumn{4}{c}{Bottleneck} & \multicolumn{4}{c}{Recurrent} \\
    $D_o^{M \rightarrow B}$ &
    33.68±11.84 & \textbf{15.83±3.80} & \textbf{10.73±4.94} & 21.86±8.88 & 33.37±11.44 & \textbf{15.56±4.61} & \textbf{11.20±6.19} & 21.93±9.05 \\
    $D_s^{M \rightarrow B}$ &
    2.35±1.02 & \textbf{0.24±0.07} & 1.75±1.10 & \textbf{0.30±0.12} & 0.64±0.14 & \textbf{0.19±0.09} & \textbf{0.25±0.16} & \textbf{0.20±0.08} \\
    $D_o^{B \rightarrow M}$ &
    31.67±13.36 & \textbf{8.01±3.62} & \textbf{9.96±7.73} & \textbf{9.14±4.47} & 35.42±15.36 & \textbf{7.56±3.07} & \textbf{10.39±7.69} & \textbf{9.08±5.10} \\
    $D_s^{B \rightarrow M}$ & 
    30.51±14.64 & 13.16±1.84 & \textbf{3.22±1.09} & 20.90±10.58 & 36.02±16.02 & 32.78±6.28 & 37.38±5.29 & 39.16±10.85 \\
    \bottomrule
    
  \end{tabular}
\end{table}

\clearpage
\subsubsection{Sensitivity analysis}
\label{append_sensitivity_analysis}
\begin{figure}[h!]
  \centering
  \includegraphics[width=1.0\textwidth]{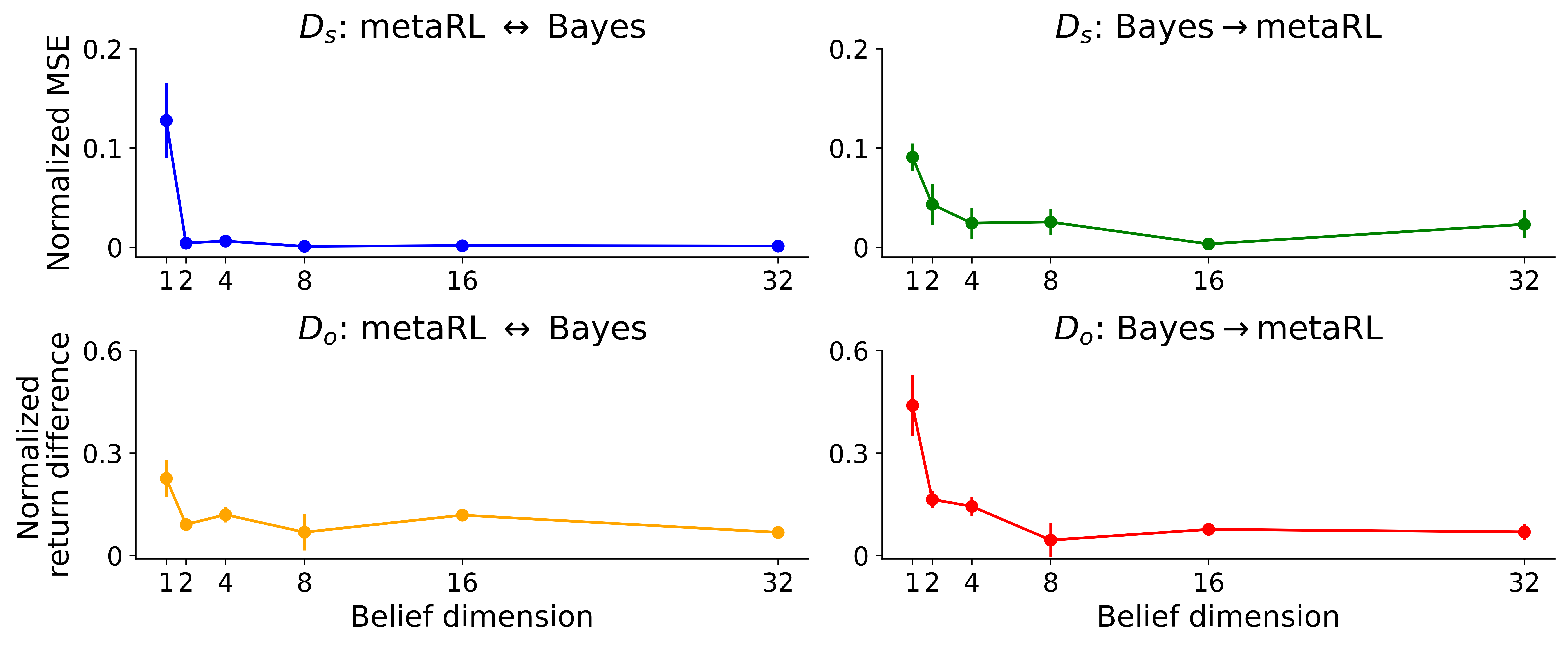}
  \caption{\textbf{Sensitivity analysis}.
  State machine simulation results of the bottleneck layers in meta-RL with predictive modules in the Oracle bandit task, as the belief dimension varies from 1 to 32.
  When the belief dimension is smaller than task complexity (e.g. <= 2), the models present with high dissimilarities in all four metrics.
  In contrast, as long as the belief dimension is greater than task complexity, all four dissimilarities remain consistently low even when the belief dimension is large, indicating that regularization is useful for maintaining compact, interpretable belief representation.
  }
  \label{fig:append_sensitivity_analysis}
\end{figure}

To evaluate whether representational equivalence between Bayes-optimal solutions and meta-RL with predictive modules is sensitive to the choice of bottleneck dimensions, using the oracle bandit task as an example, we performed systematic evaluation of state machine simulation while varying the bottleneck dimension ranging from 1 to 32. 
As shown in Fig. \ref{fig:append_sensitivity_analysis}, when the dimension of the bottleneck is greater than task complexity (i.e. the belief state dimension for the task, e.g. > 2 for the oracle bandit task), the bottleneck of the meta-RL with predictive modules achieve similarly low dissimilarities, indicating that regularization helps to maintain interpretable belief representation even when the bottleneck is more than expressive enough.
In contrast, when the dimension of the bottleneck is smaller than task complexity, both performance (suboptimal return) and representational equivalence (high dissimilarity) significantly degrade, indicating that the bottleneck dimension should be selected to be at least surpassing the task complexity in order to learn the task effectively.

\end{document}